\documentclass[10pt,twocolumn,letterpaper]{article}

\usepackage{cvpr}
\usepackage{times}
\usepackage{epsfig}
\usepackage{graphicx}
\usepackage{amsmath}
\usepackage{amssymb}

\usepackage{multirow}
\usepackage{subfigure}
\usepackage[linesnumbered,ruled]{algorithm2e}
\usepackage{amsmath}
\usepackage[breaklinks=true,bookmarks=false]{hyperref}

\cvprfinalcopy 

\def\cvprPaperID{1991} 

\ifcvprfinal\fi
\begin{document}

\title{A2-RL: Aesthetics Aware Reinforcement Learning for Image Cropping}

\author{Debang Li$^{\,1,2}$, Huikai Wu$^{\,1,2}$, Junge Zhang$^{1,2}$, Kaiqi Huang$^{1,2,3}$\\
$^{1}$ CRIPAC \& NLPR, Institute of Automation, Chinese Academy of Sciences, Beijing, China\\
$^{2}$ University of Chinese Academy of Sciences, Beijing, China\\
$^{3}$ CAS Center for Excellence in Brain Science and Intelligence Technology, Beijing, China\\
{\tt\small \{debang.li, huikai.wu,jgzhang, kaiqi.huang\}@nlpr.ia.ac.cn}
}

\maketitle
\thispagestyle{empty}

\begin{abstract}
Image cropping aims at improving the aesthetic quality of images by adjusting their composition. Most weakly supervised cropping methods (without bounding box supervision) rely on the sliding window mechanism. The sliding window mechanism requires fixed aspect ratios and limits the cropping region with arbitrary size. Moreover, the sliding window method usually produces tens of thousands of windows on the input image which is very time-consuming. Motivated by these challenges, we firstly formulate the aesthetic image cropping as a sequential decision-making process and propose a weakly supervised Aesthetics Aware Reinforcement Learning (A2-RL) framework to address this problem. Particularly, the proposed method develops an aesthetics aware reward function which especially benefits image cropping. Similar to human's decision making, we use a comprehensive state representation including both the current observation and the historical experience. We train the agent using the actor-critic architecture in an end-to-end manner. The agent is evaluated on several popular unseen cropping datasets. Experiment results show that our method achieves the state-of-the-art performance with much fewer candidate windows and much less time compared with previous weakly supervised methods.
\end{abstract}

\section{Introduction}

Image cropping is a common task in image editing, which aims to extract well-composed regions from ill-composed images. It can improve the visual quality of images, because the composition plays an important role in the image quality. An excellent automatic image cropping algorithm can give editors professional advices and help them save a lot of time~\cite{kao2017automatic}.
\begin{figure}
\begin{center}
\includegraphics[width=0.96\linewidth]{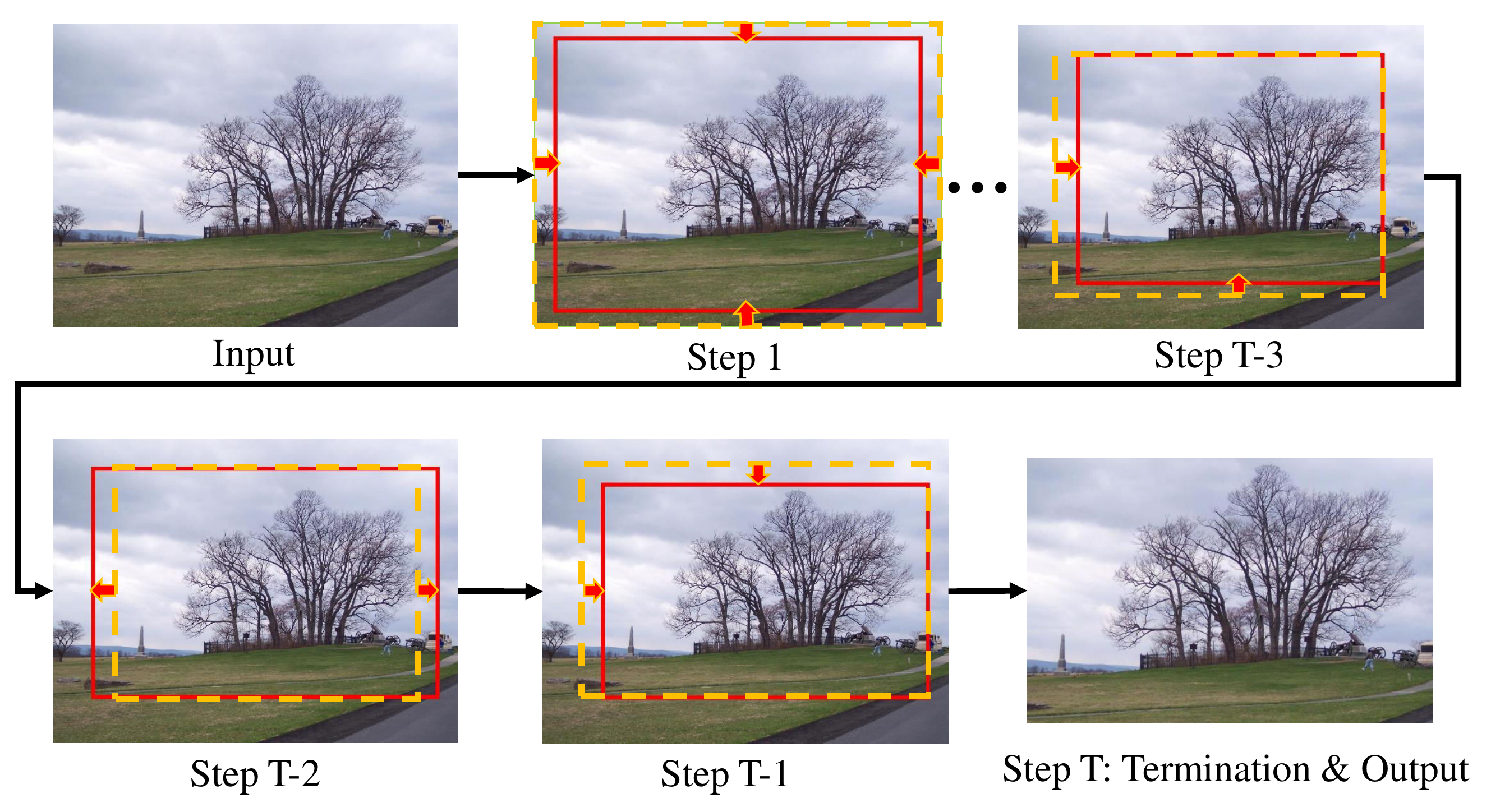}
\caption{Illustration of the sequential decision-making based automatic cropping process. The cropping agent starts from the whole image and takes actions to find the best cropping window in the input image. At each step, it takes an action (yellow and red arrow) and transforms the previous window (dashed-line yellow rectangle) to a new state (red rectangle). The agent takes the termination action and stops the cropping process to output the cropped image at step T.}
\end{center}
\label{fig:1}
\end{figure}

In the past decades, many researchers have devoted their efforts to proposing novel methods~\cite{yan2013learning,esmaeili2016fast,hongcnn} for automatic image cropping. As the cropping box annotations are expensive to obtain, several weakly supervised cropping methods (without bounding box supervision)~\cite{fang2014automatic,chen-acmmm-2017,zhang2014weakly} are proposed. Most of these weakly supervised methods follow a three-step pipeline: 1) Densely extract candidates with the sliding window method on the input image, 2) Extract carefully designed features from each region and 3) Use a classifier or ranker to grade each window and find the best region. Although these works have achieved pretty good performance, they may not find the best results due to the limitations of the sliding window method, which requires fixed aspect ratios and limits the cropping region with arbitrary size. What's more, these sliding window based methods usually need tens of thousands of candidates on image level, which is very time-consuming. Although we can set several different aspect ratios and densely extract candidates, it inevitably costs lots of time and is still unable to cover all conditions.

Based on above observations, in this paper, we firstly formulate the automatic image cropping problem as a sequential decision-making process, and propose an Aesthetics Aware Reinforcement Learning (A2-RL) model for weakly supervised cropping problem. The sequential decision-making based automatic image cropping process is illustrated in Figure~\ref{fig:1}. To our knowledge, we are the first to put forward a reinforcement learning based method for automatic image cropping. The A2-RL model can finish the cropping process within several or a dozen steps and get results of almost arbitrary shape, which can overcome the disadvantages of the sliding window method. Particularly, A2-RL model develops a novel aesthetics aware reward function which especially benefits image cropping. Inspired by human's decision making, the historical experience is also explored in the state representation to assist the current decision. We test the model on three unseen popular cropping datasets~\cite{yan2013learning,fang2014automatic,chen2017quantitative}, and the experiment results demonstrate that our method obtains the state-of-the-art cropping performance with much fewer candidate windows and much less time compared with related methods.

\section{Related Work}
Image cropping aims at improving the composition of images, which is very important for the aesthetic quality. There are a number of previous works for aesthetic quality assessment. Many early works~\cite{ke2006design,datta2006studying,luo2011content,dhar2011high} focus on designing handcrafted features based on intuitions from human's perception or photographic rules. Recently, thanks to the fast development of deep learning and newly proposed large scale datasets~\cite{murray2012ava}, there are many new works~\cite{kong2016photo,mai2016composition,deng2017image} which accomplish aesthetic quality assessment with convolutional neural networks.

Previous automatic image cropping methods can be divided into two classes, attention-based and aesthetics-based methods. The basic approach of attention-based methods~\cite{suh2003automatic,stentiford2007attention,park2012modeling,chen2016automatic} is to find the most visually salient regions in the original images. Attention-based methods can find cropping windows that draw more attention from people, but they may not generate very pleasing cropping windows, because they hardly consider about the image composition~\cite{chen2017quantitative}. For those aesthetics-based methods, they aim to find the most pleasing cropping windows from original images. Some of these works~\cite{nishiyama2009sensation,fang2014automatic} use aesthetic quality classifiers to discriminate the quality of candidate windows. Other works use RankSVM~\cite{chen2017quantitative} or RankNet~\cite{chen-acmmm-2017} to grade each candidate window. There are also change-based methods~\cite{yan2013learning}, which compares original images with cropped images so as to throw away distracting regions and retain high quality ones. Image retargeting techniques~\cite{cho2017weakly,chen2017learning} adjust the aspect ratio of an image to fit the target aspect ratio, while not discarding important content in an image, which are relevant to our task.

As for the supervision information, these methods can be divided into supervised and weakly supervised methods, depending on whether they use bounding box annotations. Supervised cropping methods~\cite{hongcnn,esmaeili2016fast,wang2015learning,wang2017deep} need bounding box annotations to train the cropper. For example, object detection based cropping methods~\cite{esmaeili2016fast,wang2017deep} are fast and effective, but they need a mount of bounding box annotations for training the detector, which is expensive. Most weakly supervised methods~\cite{fang2014automatic,chen-acmmm-2017,kao2017automatic} still rely on the sliding window method to obtain the candidate windows. As discussed above, the sliding window method uses fixed aspect ratios and limits windows with arbitrary size. What's more, these methods are also very time-consuming. In this paper, we formulate the cropping process as a sequential decision-making process and propose a weakly supervised reinforcement learning (RL) based strategy to search the cropping window. Hong~\etal~\cite{hongcnn} also regard the cropping process as a sequential process, but they use bounding box as supervision. Our RL based method can find the final results with only several or a dozen candidates of almost arbitrary size, which is much faster and more effective compared to other weakly supervised methods and doesn't need bounding box annotations compared to supervised methods.

RL based strategies have been successfully applied in many domains of computer vision, including image caption~\cite{ren2017deep}, object detection~\cite{caicedo2015active,jie2016tree} and visual relationship detection~\cite{liang2017deep}. The active object localization method~\cite{caicedo2015active} achieves the best performance among detection algorithms without region proposals. The tree-RL method~\cite{jie2016tree} uses RL to obtain region proposals and achieves comparable result with much fewer region proposals compared to RPN~\cite{renNIPS15fasterrcnn}. Above RL based object detection methods use bounding boxes as their supervision, however, our framework only uses the aesthetics information as supervision, which requires less label information. To our best knowledge, we are the first to put forward a deep reinforcement learning based method for automatic image cropping.

\section{Aesthetics Aware Reinforcement Learning}
\begin{figure*}
\begin{center}
\includegraphics[width=0.96\linewidth]{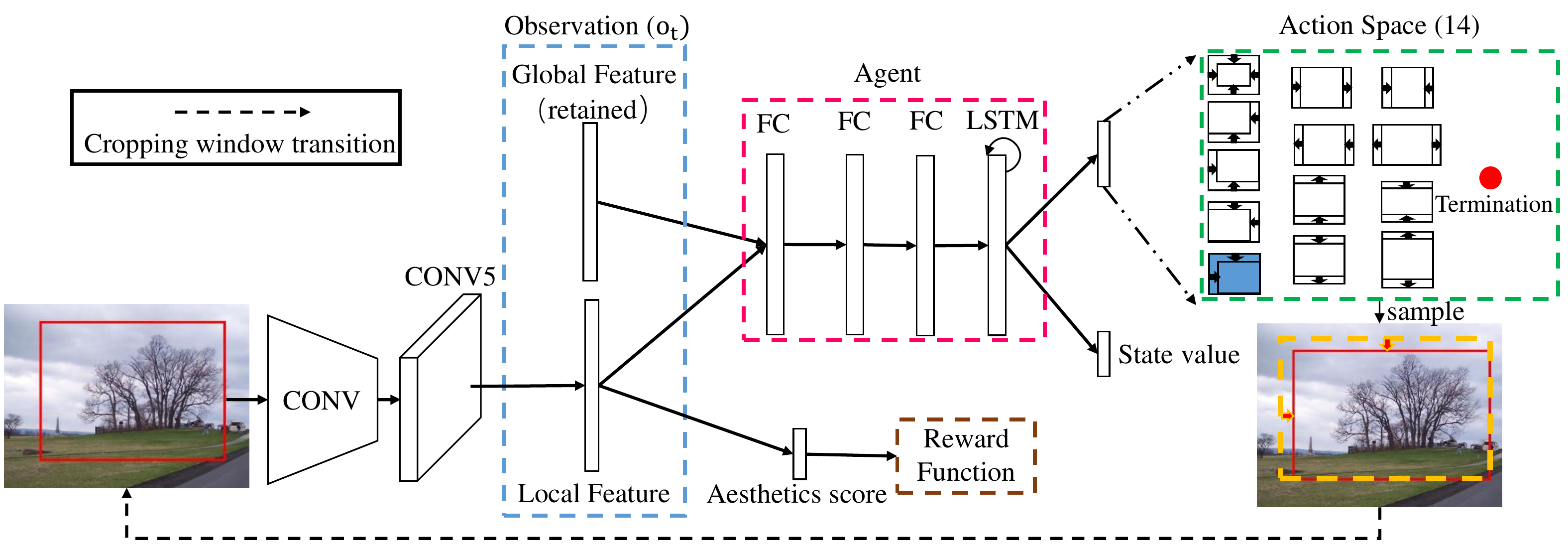}
\end{center}
\caption{Illustration of the A2-RL model architecture. In the forward pass, the feature of the cropping window (local feature) is extracted and concatenated with the feature of the whole image (global feature) which is extracted and retained previously. Then, the concatenated feature vector is fed into the actor-critic branch which has two outputs. The actor output is used to sample actions from the action space so as to manipulate the cropping window. The critic output (state value) is used to estimate the expected reward under the current state. In addition, the feature of the cropping window is also fed into the aesthetic quality assessment branch. The output of this branch is the aesthetic score of the input cropping window and stored to compute rewards for actions. In this model, both the global feature and the local feature are 1000-dim vectors, three fully-connected layers and the LSTM layer all output 1024-dim feature vectors.}

\label{fig:2}
\end{figure*}
In this paper, we formulate automatic image cropping as a sequential decision-making process. In the decision-making process, an agent interacts with the environment, and takes a series of actions to optimize a target. As illustrated in Figure~\ref{fig:2}, for our problem, the agent receives observations from the input image and the cropping window. Then it samples action from the action space according to the observation and historical experience. The agent executes the sampled action to manipulate the shape and position of the cropping window. After each action, the agent receives a reward according to the aesthetic score of the cropped image. The agent aims to find the most pleasing window in the original image by maximizing the accumulated reward. In this section, we first introduce the state space, action space and aesthetics aware reward of our model, then we detail the architecture of our aesthetics aware reinforcement learning (A2-RL) model and the training process.

\subsection{State and Action Space}
At each step, the agent decides which action to execute according to the current state. The state must provide the agent with comprehensive information for better decisions. As the A2-RL model formulates the automatic image cropping as a sequential decision-making process, the current state can be represented as $s_t=\{o_0,o_1,\cdots,o_{t-1},o_t\}$, where $o_t$ is the current observation of the agent. This formulation is similar to human's decision making process, which considers not only the current observation but also the historical experience. The historical experience is usually very valuable for future decision-making. Thus, in the proposed method, we also take the historical experience into consideration. The A2-RL model uses the features of the cropping window and the input image as the current observation $o_t$. Agent can learn about the global information and the local information with such observation. In the A2-RL model, we use a LSTM unit to memorize historical observations $\{o_0,o_1,\cdots,o_{t-1}\}$, and combine them with the current observation $o_t$ to form the state $s_t$.

We choose 14 pre-defined actions to form the action space, which can be divided into four groups: scaling actions, position translation actions, aspect ratio translation actions and a termination action. The first three groups aim to adjust the size, position and shape of the cropping window, including 5, 4 and 4 actions respectively. These three groups follow similar definitions in \cite{jie2016tree}, but with different scales. All these actions adjust the shape and position by 0.05 times of the original image size, which could capture more accurate cropping windows than a large scale. The termination action is a trigger for the agent, when this action is chosen, the agent will stop the cropping process and output the current cropping window as the final result. As the model learns when to stop the cropping process by itself, it can stop at the state where the score won't increase anymore so as to get the best cropping window. Theoretically, the agent can cover windows with almost arbitrary size and position on the original image.

The observation and action space are illustrated in Figure~\ref{fig:2} for an intuitional representation.

\subsection{Aesthetics Aware Reward}
\label{reward}
Our A2-RL model aims to find the most pleasing cropping window on the original image. So the reward function should lead the agent to find a more pleasing window at each step. We propose using the aesthetic score to evaluate the pleasing degree of images naturally. When the agent takes an action, the difference between the aesthetic scores of the new cropping window and the last one can be utilized to compute the reward for this action. More detailed, if the aesthetic score of the new window is higher than the last one, the agent will get a positive reward. On the contrary, if the score becomes lower, the agent will get a negative reward. To speed up the cropping process, we also give the agent an additional negative reward $-0.001 * (t + 1)$ at each step, where $t + 1$ is the number of steps the agent has taken since the beginning and $t$ starts from 0. This constraint will result in a lower reward when the agent takes too many steps. For an image \emph{I}, we denote its aesthetic score as $s_{\text{\emph{aes}}}(I)$. The new cropped image and the last one are denoted as $I_{t+1}$ and $I_{t}$ respectively, $sign(*)$ denote the sign function, so the foundation of our aesthetics aware reward function $r_t'$ can be formulated as :
\begin{equation}
 r_t'= sign(s_{\text{\emph{aes}}}(I_{t+1}) - s_{\text{\emph{aes}}}(I_{t})) - 0.001*(t + 1)
\label{imp}
\end{equation}
In the above definition of $r_t'$, we use the sign function to limit the variation range of $s_{\text{\emph{aes}}}(I_{t+1}) - s_{\text{\emph{aes}}}(I_{t})$, because the training is stable and easy to converge in practice under such setting. Using the reward function without the sign function makes it hard for the model to converge in our experiments, which is mainly due to the dramatic fluctuation of rewards, especially when the model samples the cropping window randomly at first.

We also consider other heuristic constraints for better cropping policies. We believe the aspect ratio of well-composed images is limited in a particular range. In the A2-RL model, if the aspect ratio of the new window is lower than 0.5 or higher than 2, the agent will receive a negative reward \emph{nr} as the penalty term for the corresponding action, so the agent can learn a strict rule not to let such situation happen. The limited range of the aspect ratio in our model is for the common cropping task, we can also modify the reward function and the action space to meet some special requirements on the aspect ratio depending on the application.  Let $ar$ denote the aspect ratio of the new window, $nr$ denote the negative reward the agent receives when the aspect ratio of the window exceeds the limited range, the whole reward function $r_t$ for the agent taking an action $a_t$ under the state $s_t$ can be formulated as:
\begin{equation}
r_t(s_t, a_t) = \left\{\begin{array}{cc}
r_t' + \emph{nr},&\mbox{if $ar<0.5$ or  $ar>2$}\\
r_t',&\mbox{otherwise}
\end{array}
\right.
\label{r}
\end{equation}
\subsection{A2-RL Model}
With the defined state space, action space and reward function, we start to introduce the architecture of our Aesthetics Aware Reinforcement Learning (A2-RL) framework. The detailed architecture of the framework is illustrated in Figure~\ref{fig:2}. The A2-RL model starts with a 5-layer convolution block and a fully-connected layer which outputs 1000-dimensional vector for feature representation. Then the model splits into two branches, the first one is the actor-critic branch, the other is the aesthetic quality assessment branch. The actor-critic branch is composed of three fully-connected layers and a LSTM layer. The LSTM layer is used to memorize the historical observations. The actor-critic branch has two outputs, the first one is the policy output, which is also named \textbf{Actor}, the other output is the value output, also named \textbf{Critic}. The policy output is a fourteen-dimensional vector, each dimension corresponding to the probability of taking relevant action. The value output is the estimation of the current state, which is the expected accumulated reward in the current situation. The aesthetic quality assessment branch outputs an aesthetic quality score for the cropped image, which is used to compute the reward.

In the image cropping process, the A2-RL model provides the agent with the probability of each action under the current state. As shown in Figure~\ref{fig:2}, the model feeds the cropped image into the feature representation unit and extracts the local feature at first. Then the feature is combined with the global feature which is extracted in the first forward pass and retained for the following process. The combined feature vector is then fed into the actor-critic branch. According to the policy output, the agent samples the relevant action and adjusts the size and position of the cropping window correspondingly. For example, in Figure~\ref{fig:2}, the agent executes the sampled action to shrink the cropping window from left and top with 0.05 times the size of the image. Forward pass will continue until the termination action is sampled.
\subsection{Training A2-RL Model}
In the A2-RL, we propose using the asynchronous advantage actor-critic (A3C) algorithm~\cite{mnih2016asynchronous} to train the cropping policy.  Different from the original A3C, we replace the asynchronous mechanism with mini-batch to increase the diversity. In the training stage, we use the advantage function~\cite{mnih2016asynchronous} and entropy regularization term~\cite{williams1991function} to form the optimization objective of the policy output. We use $R_t$ to denote the accumulated reward at step $t$, which is $\sum_{i=0}^{k-1}\gamma^ir_{t+i} + \gamma^kV(s_{t+k};\theta_v)$, where $\gamma$ is the discount factor, $r_t$ is the aesthetics aware reward at step t, $V(s_t;\theta_v)$ is the value output under state $s_t$, $\theta_v$ denotes the network parameters of \textbf{Critic} branch and $k$ ranges from 0 to $t_{max}$. $t_{max}$ is the maximum number of steps before updating. The optimization objective of the policy output is to maximize the advantage function $R_t-V(s_t;\theta_v)$ and the entropy of the policy output $H(\pi(s_t;\theta))$, where $\pi(s_t;\theta)$ is the probability distribution of policy output, $\theta$ denotes the network parameters of \textbf{Actor} branch, and $H(*)$ is the entropy function. The entropy in the optimization objective is used to increase the diversity of actions, which can make the agent learn flexible policies. The optimization objective of the value output is to minimize $(R_t -V(s_t;\theta_v))^2/2$. So gradients of the actor-critic branch can be formulated as $\nabla_{\theta}log \pi(a_t|s_t;\theta)(R_t-V(s_t;\theta_v))+\beta\nabla_{\theta} H(\pi(s_t;\theta))$ and $\nabla_{\theta_v}(R_t - V(s_t;\theta_v))^2/2$, where $\beta$ is to control the influence of entropy and $\pi(a_t|s_t;\theta)$ is the probability of the sampled action $a_t$ under the state $s_t$.
\begin{algorithm}
\caption{Training procedure of the A2-RL model}
\label{algo:A2-RL Training}
\KwIn{original image \emph{I}}

$f_{global} = Feature\_extractor(\emph{I})$

$I_0\leftarrow I$, $t\leftarrow 0$


\Repeat{$t == T_{max}$ or $a_{t-1}$ is termination action}{

$t_{start}=t$, $d\theta\leftarrow0$, $d\theta_v\leftarrow0$

\Repeat{$t-t_{start} == t_{max}$ or $a_{t-1}$ is termination action}{
$f_{local} = Feature\_extractor(\emph{$I_t$})$

$o_t = concat(f_{global}, f_{local})$

$s_t = LSTM_{AC}(o_t)$ //LSTM of Actor-Critic

Perform $a_t$ according to the policy output $\pi(a_t|s_t;\theta)$ and get the new image $I_{t+1}$

$r_t = reward(I_t, I_{t+1}, t)$

$t = t+1$
}
$R=\left\{\begin{array}{ll}0 &\mbox{if $a_{t-1}$ is termination action}\\ V(s_t;\theta_v) &\mbox{for other actions}\end{array}\right.$

\For{$i\in\{t-1,...,t_{start}\}$}{
$R\gets r_i + \gamma R$

$d\theta\gets d\theta+\nabla_{\theta}log \pi(a_i|s_i;\theta)(R-V(s_i;\theta_v))$
$    \qquad+\beta\nabla_{\theta} H(\pi(s_i;\theta))$

$d\theta_v\gets d\theta_v + \nabla_{\theta_v}(R - V(s_i;\theta_v))^2/2$
}
Update $\theta$ with $d\theta$ and $\theta_v$ with $d\theta_v$
}
\end{algorithm}

The whole training procedure of the A2-RL model is described in Algorithm~\ref{algo:A2-RL Training}. $T_{max}$ means maximum number of steps the agent takes before termination.

\section{Experiments}
\subsection{Experimental Settings}
\paragraph{Training Data}
To train our network, we select images from a large scale aesthetics image dataset named AVA~\cite{murray2012ava}, which consists of $\sim$250000 images. All these images are labeled with aesthetic score rating from one to ten by several people. As the score distribution is extremely unbalanced, we simply divide them into three classes: low quality, middle quality and high quality. These three classes correspond to score from one to four, four to seven and seven to ten respectively. We choose about 3000 images from each class to compose the training set. Finally, there are $\sim$9000 images in the training set. With these training data, our model can learn policies with images of diverse quality, which can make the model generalize well to different images.
\paragraph{Implementation Details}
In our experiment, the aesthetic score $s_{\text{\emph{aes}}}(\emph{I})$ of the image \emph{I} is the output of the pre-trained view finding network (VFN) \cite{chen-acmmm-2017}, which is an aesthetic ranker modified from the original AlexNet \cite{krizhevsky2012imagenet}. The VFN is trained with the same training data and ranking loss as the original settings~\cite{chen-acmmm-2017}. As shown in Figure~\ref{fig:2}, the actor-critic branch share the feature extractor unit with the VFN.

RMSProp~\cite{tieleman2012lecture} algorithm is utilized to optimize the A2-RL model, the learning rate is set to 0.0005 and the other arguments are set by default values. The mini batch size in training is set to 32. The discount factor $\gamma$ is set as 0.99 and the weight of entropy loss $\beta$ is set as 0.05 respectively. The $T_{max}$ is set as 50, and the update period $t_{max}$ is set to 10. The penalty term \emph{nr} in reward function is empirically set to -5, which can lead to a strict rule that prevents the aspect ratio of the cropping window exceeding the limited range.

We also choose 900 images from AVA dataset as the validation set following the way of the training set. As the A2-RL model aims to find the cropping window with the highest aesthetic score, on the validation set, we use the improvement of aesthetic score between the original and cropped images as metric. We train the networks on the training set for 20 epochs and validate the models on the validation set every epoch. The model which achieves the best average improvement on the validation set is chosen as the final model.
\paragraph{Evaluation Data and Metrics}
\begin{table}
\centering
\begin{tabular}{c|c|c}\hline
Method&Avg IoU&Avg Disp Error\\\hline
eDN \cite{vig2014large}&0.4857&0.1372\\
RankSVM+DeCAF$_7$~\cite{chen2017quantitative}&0.6019&0.1060\\
VFN+SW~\cite{chen-acmmm-2017}&0.6328&0.0982\\\hline
A2-RL w/o \emph{nr}&0.5720&0.1178\\
A2-RL w/o LSTM&0.6310&0.1014\\
A2-RL(Ours)&\textbf{0.6633}&\textbf{0.0892}\\\hline
\end{tabular}

\centering
\caption{Cropping Accuracy on FCD~\cite{chen2017quantitative}.}
\label{FCDB}
\end{table}
\begin{table*}[t]
\centering
\begin{tabular}{c|c|c|c|c|c|c}\hline
\multirow{2}{*}{Method}&\multicolumn{2}{c|}{Annotation I}&\multicolumn{2}{c|}{Annotation II}&\multicolumn{2}{c}{Annotation III}\\
\cline{2-7}&Avg IoU&Avg Disp Error&Avg IoU&Avg Disp Error&Avg IoU&Avg Disp Error\\\hline
eDN \cite{vig2014large}&0.4636&0.1578&0.4399&0.1651&0.4370&0.1659\\
RankSVM+DeCAF$_7$ \cite{chen2017quantitative}&0.6643&0.092&0.6556&0.095&0.6439&0.099\\
LearnChange \cite{yan2013learning}&0.7487&0.0667&0.7288&0.0720&0.7322&0.0719\\
VFN+SW \cite{chen-acmmm-2017}&0.7401&0.0693&0.7187&0.0762&0.7132&0.0772\\\hline
A2-RL w/o \emph{nr}&0.6841&0.0852&0.6733&0.0895&0.6687&0.0895\\
A2-RL w/o LSTM&0.7855&0.0569&0.7847&0.0578&0.7711&0.0578\\
A2-RL(Ours)&\textbf{0.8019}&\textbf{0.0524}&\textbf{0.7961}&\textbf{0.0535}&\textbf{0.7902}&\textbf{0.0535}\\\hline
\end{tabular}

\caption{Cropping Accuracy on CUHK-ICD~\cite{yan2013learning}.}
\label{ICDB}
\end{table*}
To evaluate the capacities of our agent, we test it on three unseen automatic image cropping datasets, including CUHK Image Cropping Dataset (CUHK-ICD)~\cite{yan2013learning}, Flickr Cropping Dataset (FCD)~\cite{chen2017quantitative} and Human Cropping Dataset (HCD)~\cite{fang2014automatic}. The first two datasets use the same evaluation metrics, while the last one uses different metrics. We adopt the same metrics as the original works for fair comparison.

There are 950 test images in CUHK-ICD, which are annotated by three different expert photographers. FCD contains 348 test images, and each image has only one annotation. On these two datasets, previous works~\cite{yan2013learning,chen2017quantitative,chen-acmmm-2017} use the same evaluation metrics to measure the cropping accuracy, including \emph{average intersection-over-union (IoU)} and \emph{average boundary displacement}. In this paper, we denote the ground truth window of the image \emph{i} as $W_i^g$ and the cropping window as $W_i^c$. The average \emph{IoU} of N images can be computed as
\begin{equation}
1/N\sum_{i=1}^Narea(W_i^g\cap W_i^c)/area(W_i^g\cup W_i^c)
\end{equation}
The average boundary displacement computes the average distance between the four edges of the ground truth window and the cropping window. In image \emph{i}, we denote four edges of the ground truth window as $B_i^g(l)$, $B_i^g(r)$, $B_i^g(u)$, $B_i^g(b)$, correspondingly, four edges of the cropping window are denoted as $B_i^c(l)$, $B_i^c(r)$, $B_i^c(u)$, $B_i^c(b)$. The \emph{average boundary displacement} of N images can be computed as
\begin{equation}
1/N\sum_{i=1}^N\sum_{j=\{l,r,u,b\}}|B_i^g(j)-B_i^c(j)|/4
\end{equation}

HCD contains 500 test images, each is annotated by ten people. Because it has more annotations for each image than the first two datasets, the evaluation metric is a little different. Previous works~\cite{fang2014automatic,kao2017automatic} on this dataset use \emph{top-K maximum IoU} as the evaluation metric, which is similar to the previous average IoU. \emph{Top-K maximum IoU} metric computes the IoU between the proposed cropping windows and ten ground truth windows, then it chooses the maximum IoU as the final result. \emph{Top-k} means to use \emph{k} best cropping windows to compute the result.

\subsection{Evaluation of Cropping Accuracy}
In this section, we compare the cropping accuracy of our A2-RL model with previous sliding window based weakly supervised methods to validate its effectiveness. As the aesthetic assessment of our model is based on VFN~\cite{chen-acmmm-2017}, we mainly compare our model with this method. Our model uses RL based method to search the best cropping windows sequentially with only several candidates. The VFN-based method uses sliding window to densely extract candidates. We also compare with several other baselines.
\paragraph{Cropping Accuracy on CUHK-ICD and FCD}
As the previous VFN method~\cite{chen-acmmm-2017} is only evaluated on CUHK-ICD~\cite{yan2013learning} and FCD~\cite{chen2017quantitative}, we also mainly compare our framework with VFN on these two datasets. Notably, the original VFN not only uses the sliding window candidates, but also uses the \emph{ground truth} window of test images as candidates, which leads to a remarkably high performance on these two datasets. As A2-RL model aims to search the best cropping window, and in practice, there won't be any ground truth window for cropping algorithms, so, in this experiment, we don't use any ground truth windows in both frameworks for fair comparison. It's also worthy to mention that, the A2-RL model has never seen images from both datasets during training.

Besides the two frameworks discussed above, we also compare some other cropping methods. We choose the best attention-based method eDN reported in \cite{chen2017quantitative} on behalf of the attention-based cropping algorithms. This method computes the saliency maps with algorithms from \cite{vig2014large}, and search the best cropping window by maximizing the difference of average saliency between the cropping window and other region. We also choose the best result (\emph{RankSVM+DeCAF$_7$}) reported in \cite{chen2017quantitative} as another baseline. In this method, aesthetic feature \emph{DeCAF$_7$} is extracted from AlexNet and a \emph{RankSVM} is trained to find the best cropping window among all the candidates. For all these sliding window based methods, including \emph{eDN, RankSVM+DeCAF$_7$} and \emph{VFN+SW (sliding window)}, the results are all reported with the same sliding window setting as \cite{chen2017quantitative}.

Experiments on FCD are shown in Table~\ref{FCDB}, where \emph{VFN+SW} and \emph{A2-RL} are the two mainly comparable frameworks. We also show the results on CUHK-ICD in Table~\ref{ICDB}. As there are 3 annotations for each image, following previous works~\cite{yan2013learning,chen2017quantitative,chen-acmmm-2017}, we list the results for each annotation separately. All symbols in Table~\ref{ICDB} are the same as Table~\ref{FCDB}. What's more, we also report the best result in \cite{yan2013learning}, in which this dataset is proposed. Notably, the method is trained with supervised cropping data on this dataset, which is not very fair for us to compare. As this method is change-based, we denote it as \emph{LearnChange} in Table~\ref{ICDB}.

From Tables~\ref{FCDB} and \ref{ICDB}, we can see that our A2-RL model outperforms other methods consistently on these two datasets, which demonstrates the effectiveness of our model.

\paragraph{Cropping Accuracy on HCD}
We also evaluate our A2-RL model on HCD~\cite{fang2014automatic}. Following previous works~\cite{fang2014automatic,kao2017automatic} on this dataset, \emph{top-K maximum IoU} is employed as the metric of cropping accuracy. We choose two state-of-the-art methods~\cite{fang2014automatic,kao2017automatic} on this dataset as our baselines. The results are shown in Table~\ref{HCDB}. As our A2-RL model finds one cropping window at a time, we compare the results using the \emph{top-1 Max IoU} as metric. From Table~\ref{HCDB}, we can see that our A2-RL model still outperforms the state-of-the-art methods.
\begin{table}
\centering
\begin{tabular}{c|c}\hline
Method&Top-1 Max IoU\\\hline
Fang \etal \cite{fang2014automatic}&0.6998\\
Kao \etal \cite{kao2017automatic}&0.7500\\\hline
A2-RL w/o \emph{nr}&0.7089\\
A2-RL w/o LSTM&0.7960\\
A2-RL(Ours)&\textbf{0.8204}\\\hline
\end{tabular}
\caption{Cropping Accuracy on HCD~\cite{fang2014automatic}.}
\label{HCDB}
\end{table}

\subsection{Evaluation of Time Efficiency}

In this section, we study the time efficiency of our A2-RL model. We compare our model with the sliding window based VFN model on FCD. Experimental results are shown in Table~\ref{Time}. The \emph{Avg Steps} and \emph{Avg Time} mean the average value of steps and time methods cost to finish the cropping process on a single image. We also augment the number of sliding windows in this experiment. Notably, all results in Table~\ref{Time} are evaluated on the same machine, which has a single NVIDIA GeForce Titan X pascal GPU with 12GB memory and Intel Core i7-6800k CPU with 6 cores.

\begin{table}[t]
\centering
\begin{tabular}{c|c|c|c|c}\hline
\multirow{2}{*}{Method}&\multicolumn{1}{c|}{Avg}&\multicolumn{1}{c|}{Avg}&\multicolumn{1}{c|}{Avg}&\multicolumn{1}{c}{Avg}\\
{}&IoU&Disp&Steps&Time(s)\\\hline
VFN+SW&0.6328&0.0982&137&1.29\\
VFN+SW+&0.6395&0.0956&500&4.37\\
VFN+SW++&0.6442&0.0938&1125&9.74\\\hline
A2-RL(Ours)&\textbf{0.6633}&\textbf{0.0892}&\textbf{13.56}&\textbf{0.245}\\\hline
\end{tabular}
\centering
\caption{Time Efficiency comparison on FCD~\cite{chen2017quantitative}. VFN+SW, VFN+SW+ and VFN+SW++ correspond different number of candidate windows, where VFN+SW follows original setting \cite{chen-acmmm-2017}.}
\label{Time}
\end{table}

From Table~\ref{Time}, we can easily find that the cropping accuracy is improved as we augment the number of sliding windows, but the consumed time also grows. Unsurprisingly, our A2-RL model needs much fewer steps and costs much less time than other methods. The average number of steps our A2-RL model takes is more than 10 times less than the sliding window based methods, but our A2-RL model still gets better cropping accuracy. These results show the capacities of our RL-based model, with the novel aesthetics aware reward and history-preserved state representation, our model learns to use as few actions as possible to obtain a more pleasant image.

\subsection{Experiment Analysis}
In this section, we analyse the experiment results and study our model.
\paragraph{RL Search vs. Sliding Window}
From Tables~\ref{FCDB}, \ref{ICDB} and \ref{Time}, we can find out that the \emph{A2-RL} method is better than the \emph{VFN+SW} method in cropping accuracy and time efficiency consistently. The main difference between these two methods is the way to get the cropping candidates. From this observation, we conclude that our proposed RL-based search method is better than the sliding window method, which is very obvious. Although the sliding window method can densely extract candidates, it still fails to find very accurate candidates due to the fixed aspect ratios. On the contrary, our A2-RL model can find cropping windows with almost arbitrary size.
\paragraph{Observation+History Experience vs. only Observation}
We use LSTM unit to memorize historical observations $\{o_0,o_1,\cdots,o_{t-1}\}$ and combine them with the current observation $o_t$ to form the state $s_t$. In this section, we study the effect of the history experience in our model. We abandon the LSTM unit in the A2-RL model, so the agent only uses the current observation $o_t$ as the state $s_t$ to make decisions. We train a new agent under such setting and evaluate it on above three datasets. Results are shown in Tables~\ref{FCDB}, \ref{ICDB} and \ref{HCDB}, where the new agent is denoted as \emph{A2-RL w/o LSTM}. From these results, we can find that the cropping accuracy of the new model is much lower than the original A2-RL model, which demonstrates the importance of historical experiences.
\paragraph{The effect of the limited aspect ratio.}
As shown in Equation~\ref{r}, if the aspect ratio of the cropped image exceeds the limited range, the agent will get an additional negative reward \emph{nr}. In this section, we study the effect of the penalty term \emph{nr} in the reward function. We remove the penalty term \emph{nr} in the reward function and train a new agent. The new agent is evaluated on the above three datasets and the results are shown in Tables~\ref{FCDB}, \ref{ICDB} and \ref{HCDB}, where the new agent is denoted as \emph{A2-RL w/o nr}. From these results, we can find that the cropping accuracy of the new agent also decreases a lot, which demonstrates the importance of the penalty term \emph{nr} in the reward function.
\begin{figure*}[ht]
\begin{center}
\includegraphics[width=0.9\linewidth]{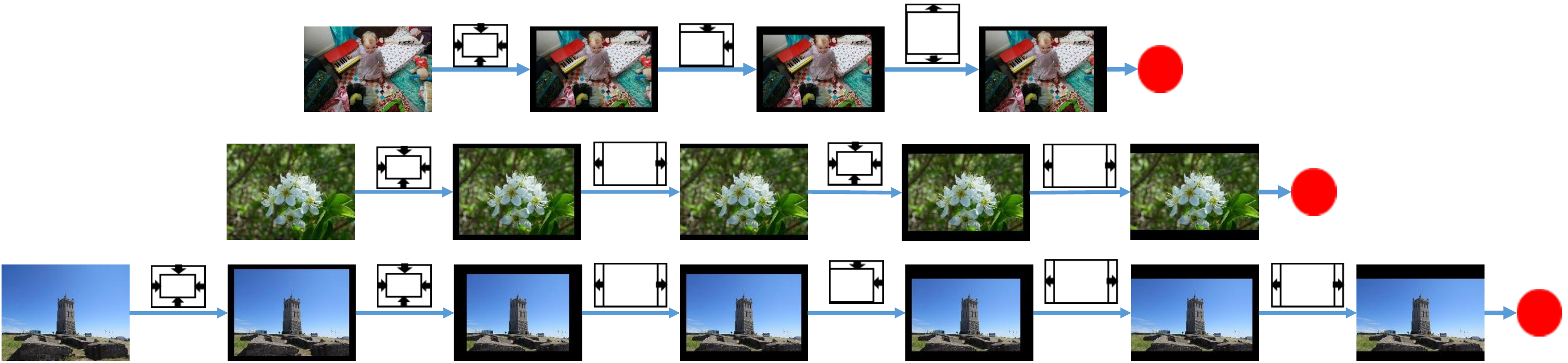}
\end{center}
\caption{Examples of the sequential actions selected by the agent and corresponding intermediate results. Images are from FCD~\cite{chen2017quantitative}.}
\label{fig:3}
\end{figure*}

\begin{figure*}[ht]
\begin{center}
\subfigure[Input Image]{
\centering
\begin{minipage}[b]{0.14\textwidth}
\includegraphics[width=1\textwidth]{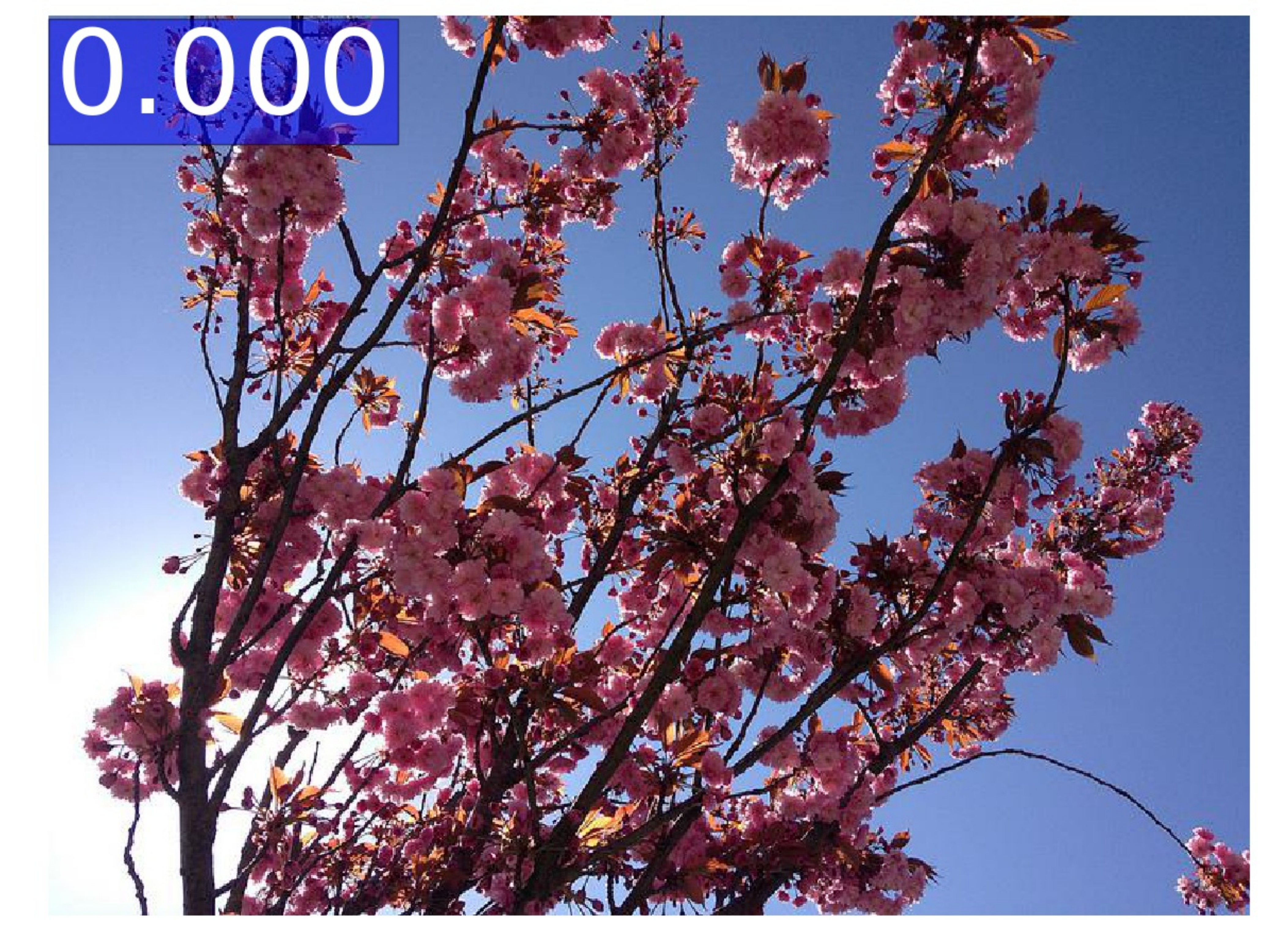} \\
\includegraphics[width=1\textwidth]{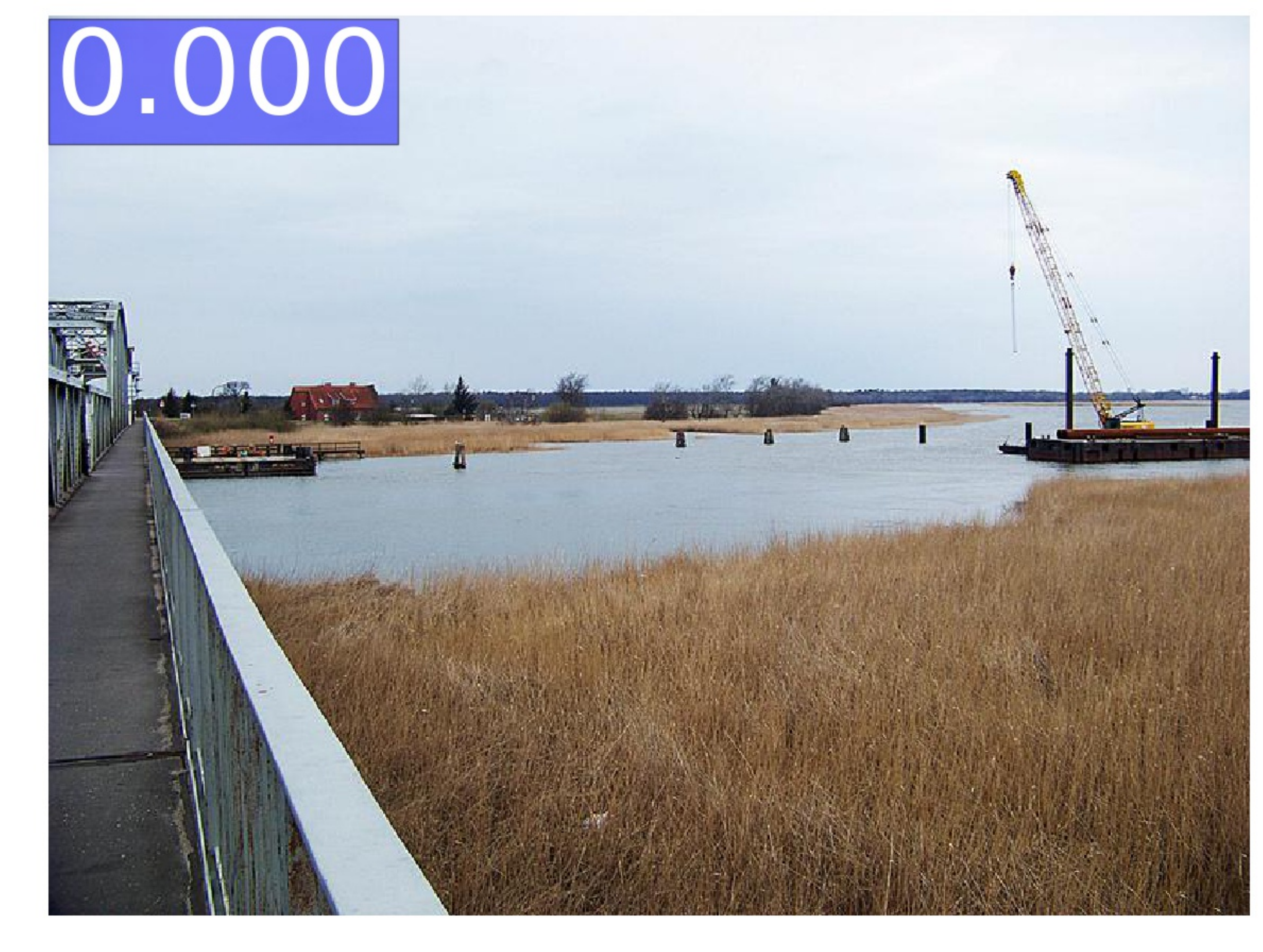} \\
\includegraphics[width=1\textwidth]{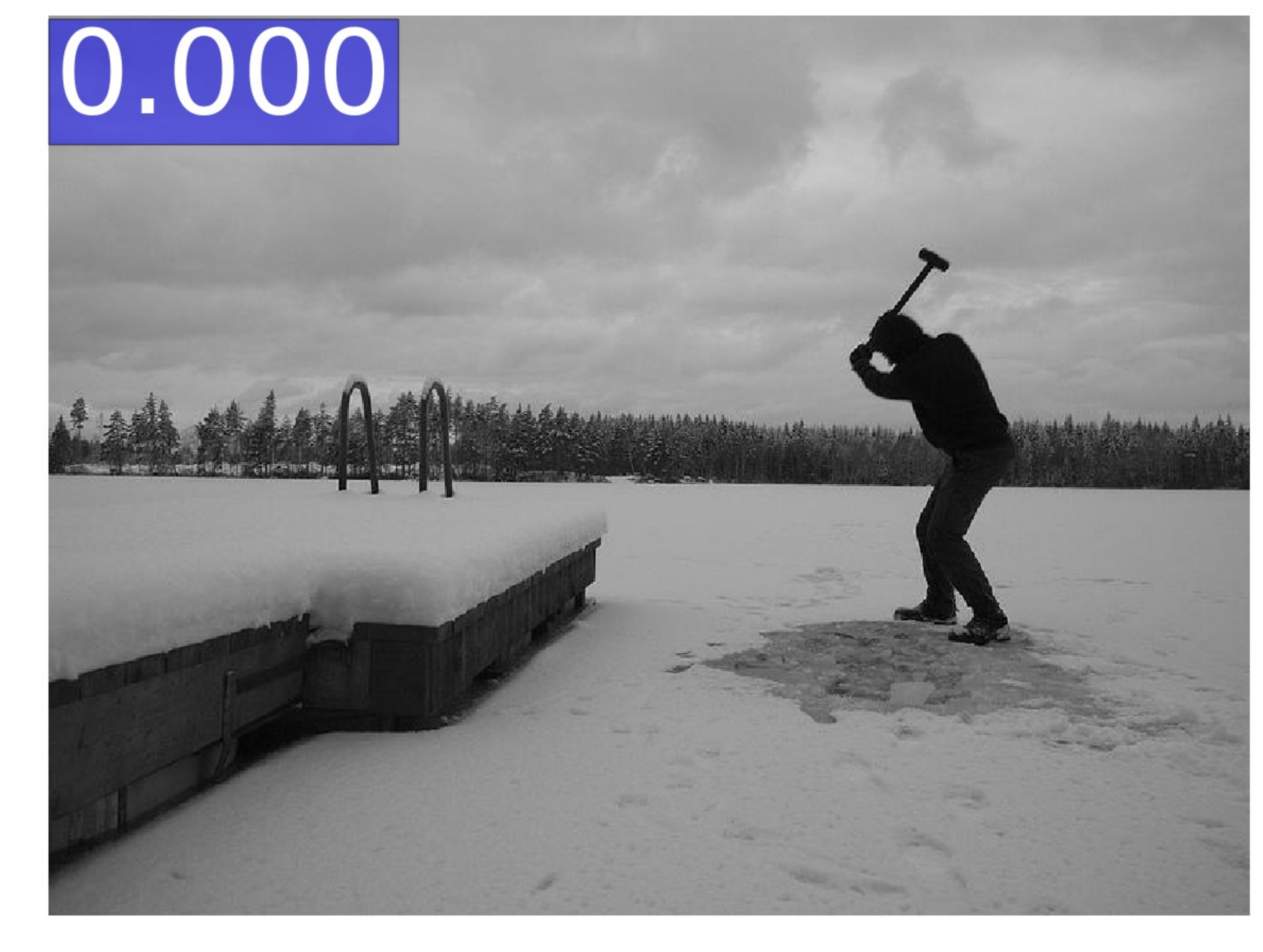}\\
\includegraphics[width=1\textwidth]{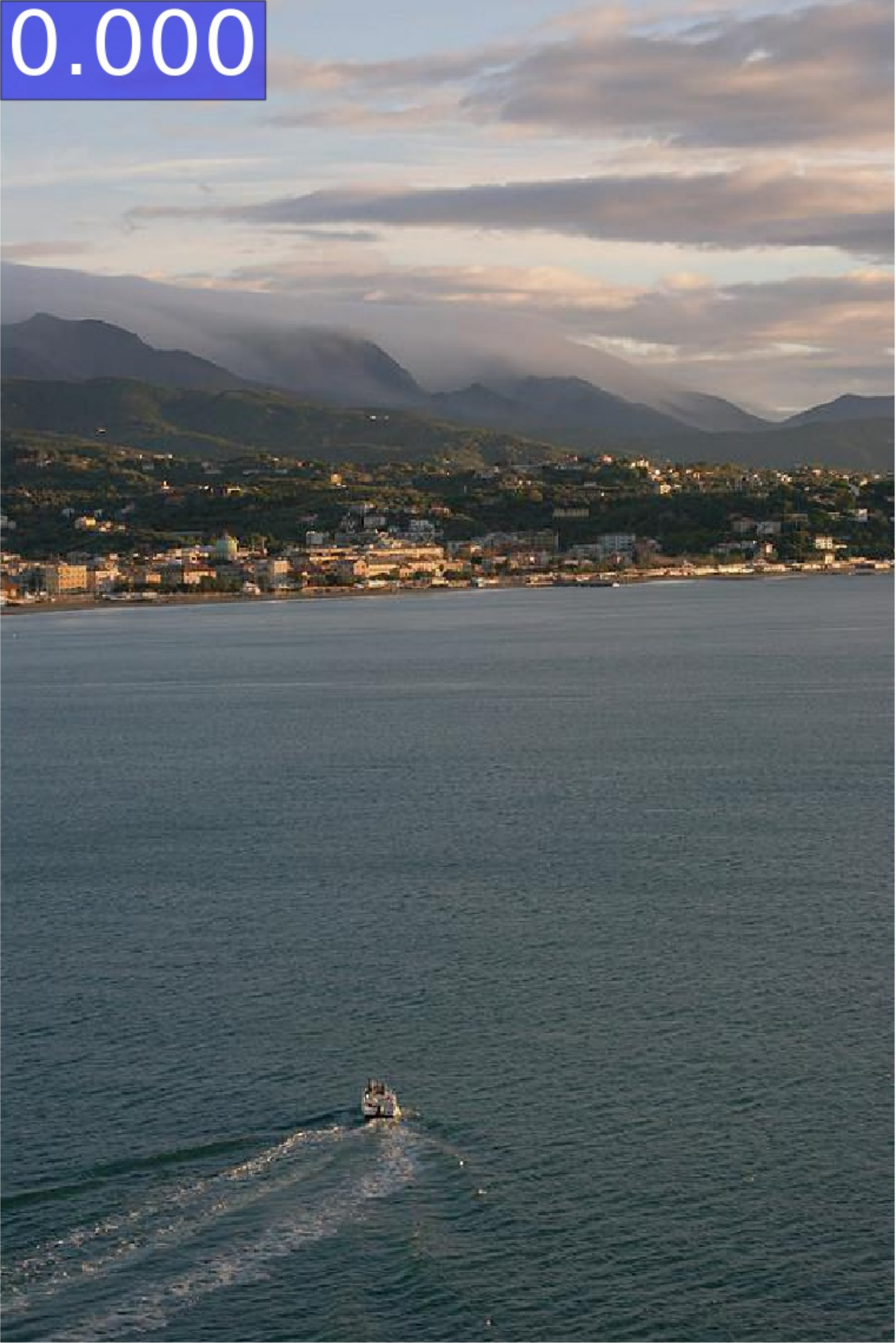}
\end{minipage}
}
\subfigure[VFN+SW~\cite{chen-acmmm-2017}]{
\centering
\begin{minipage}[b]{0.14\textwidth}
\includegraphics[width=1\textwidth]{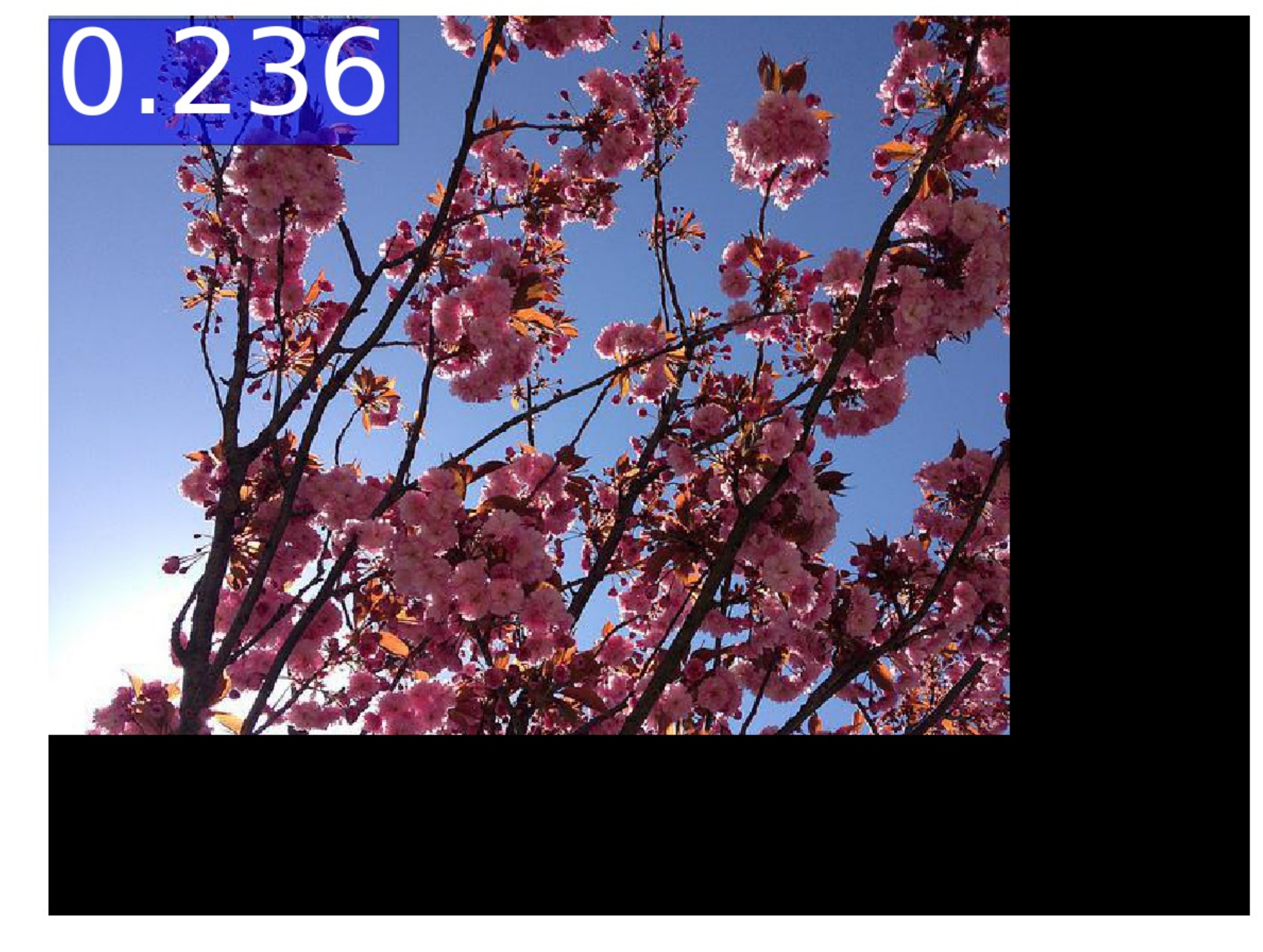} \\
\includegraphics[width=1\textwidth]{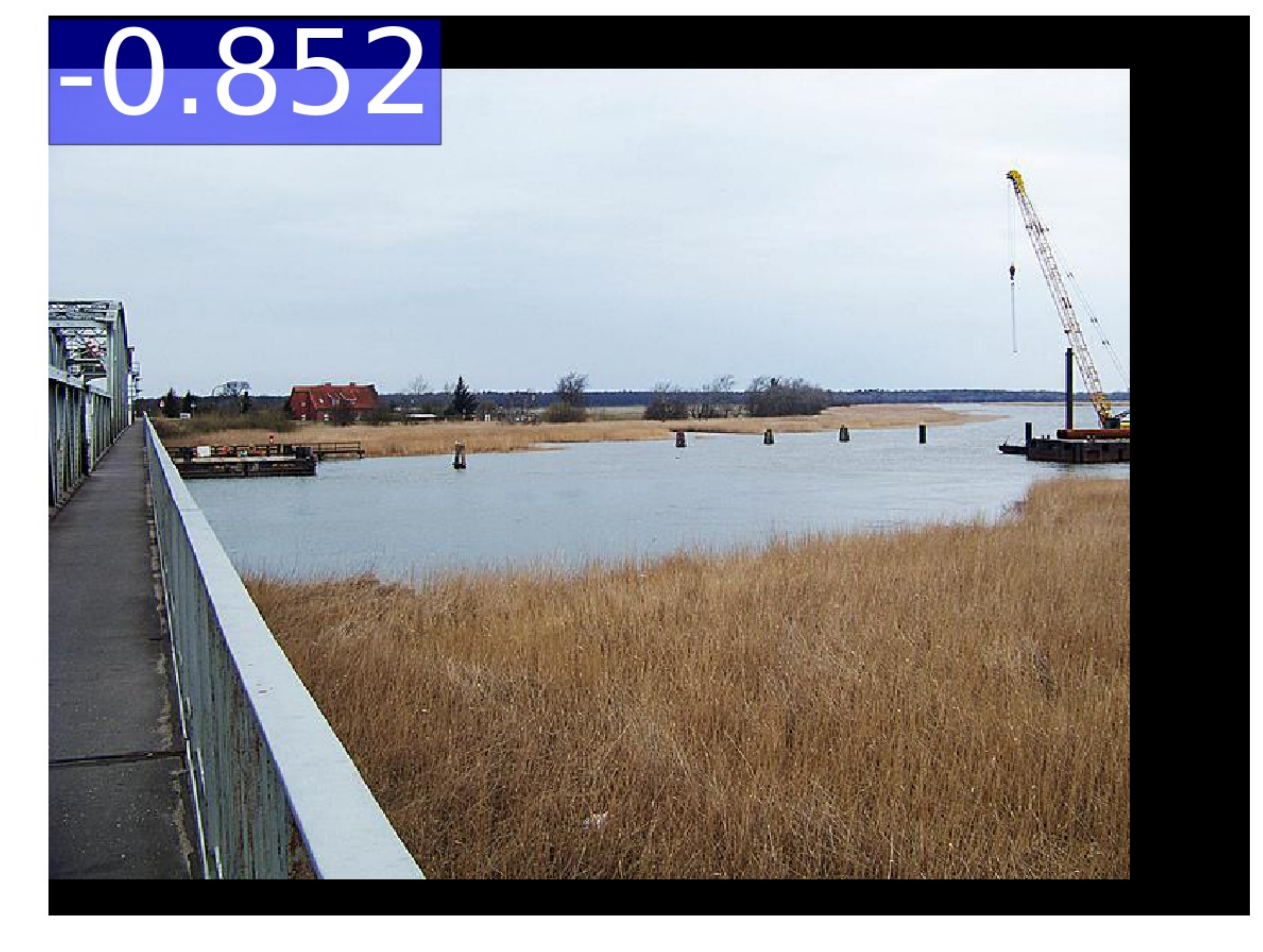} \\
\includegraphics[width=1\textwidth]{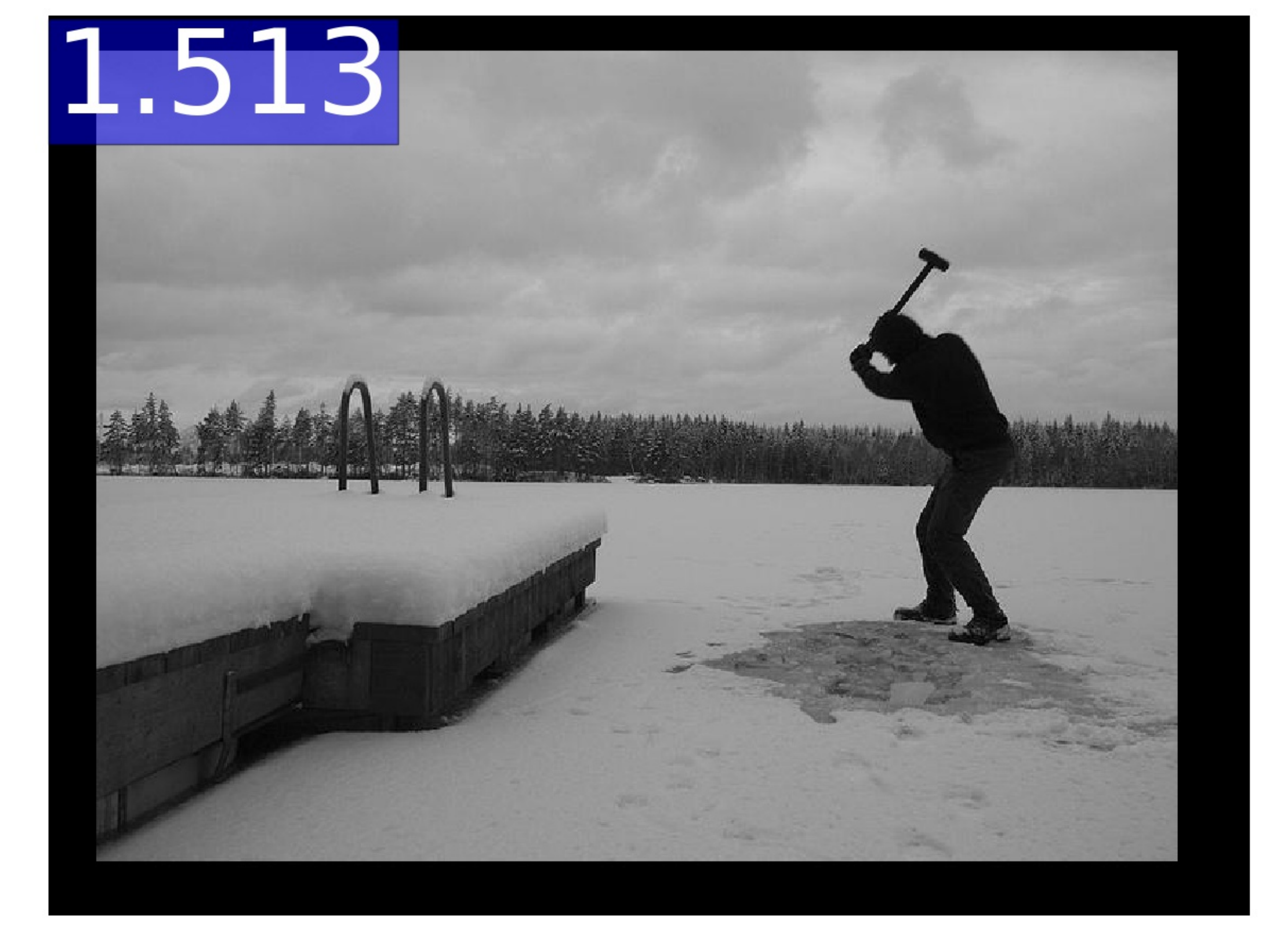}\\
\includegraphics[width=1\textwidth]{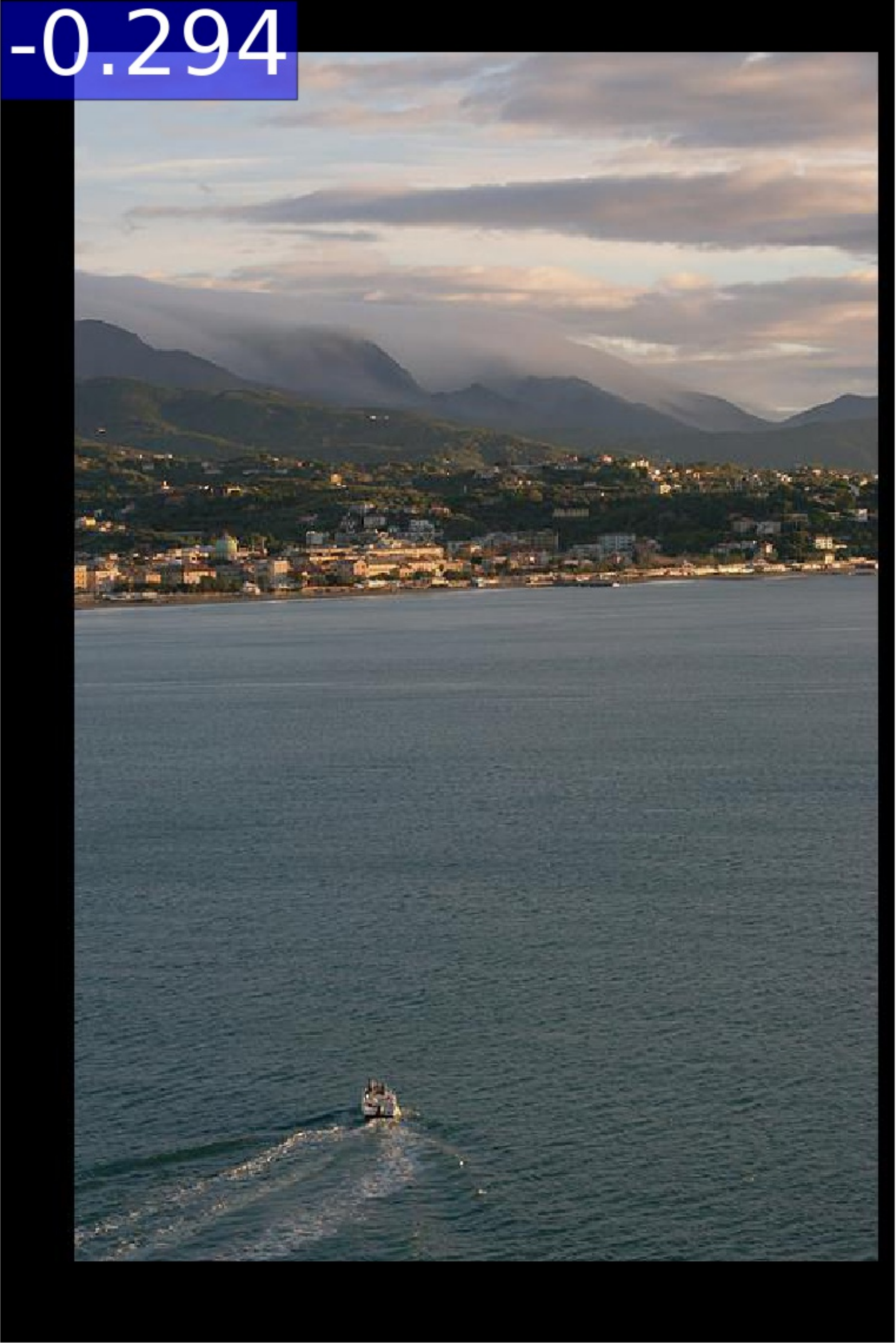}
\end{minipage}
}
\subfigure[A2-RL w/o \emph{nr}]{
\centering
\begin{minipage}[b]{0.14\textwidth}
\includegraphics[width=1\textwidth]{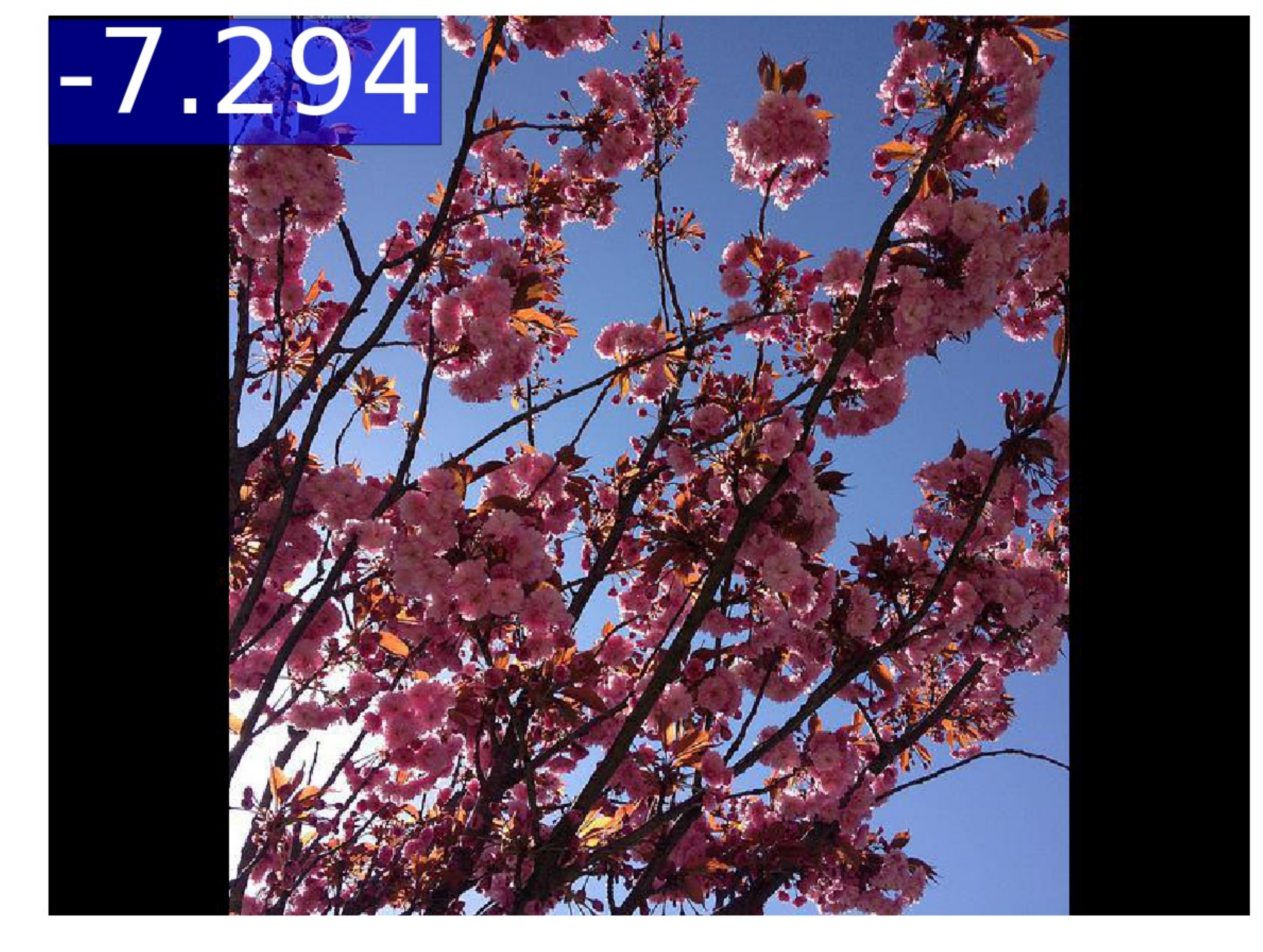} \\
\includegraphics[width=1\textwidth]{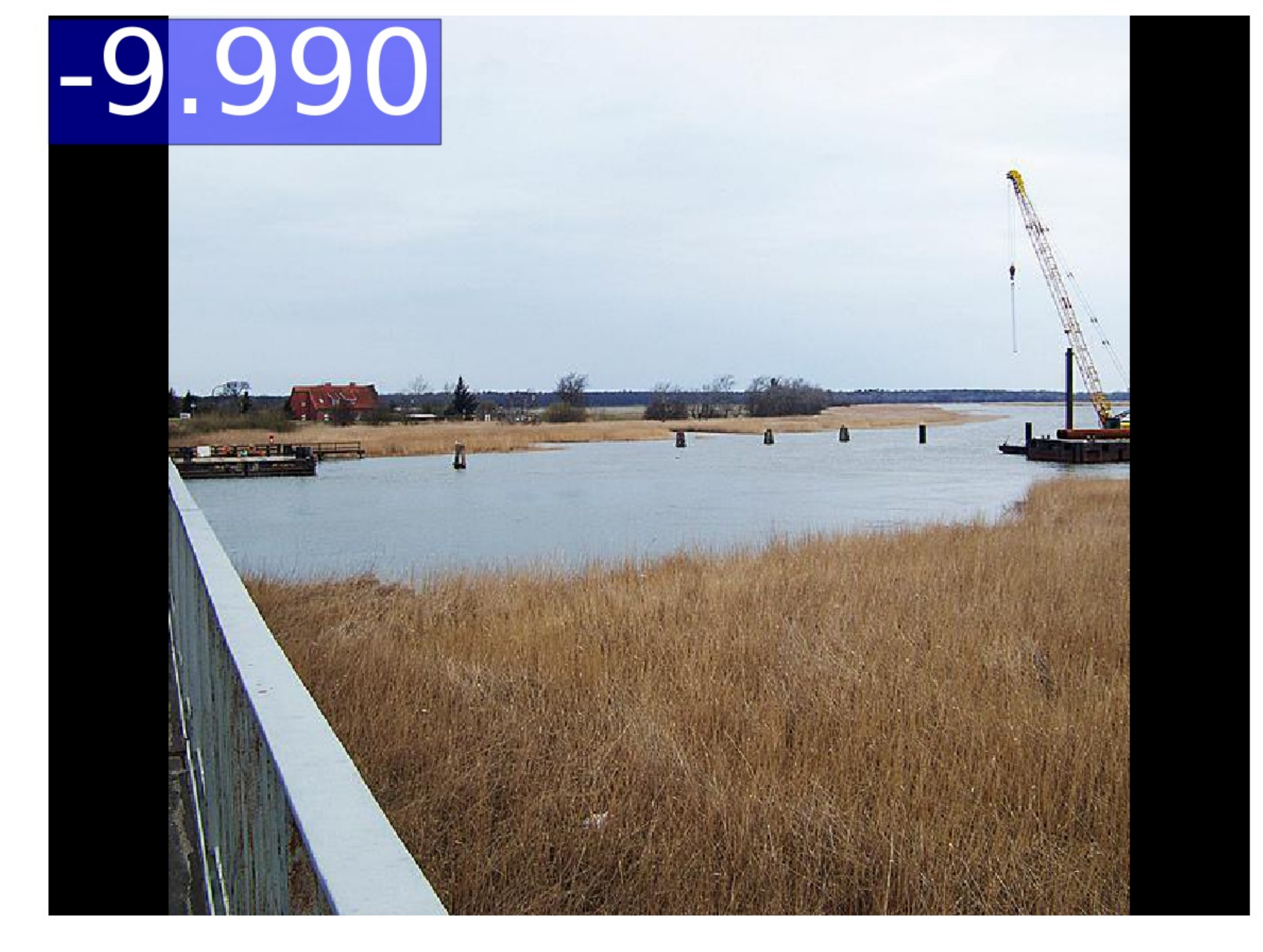} \\
\includegraphics[width=1\textwidth]{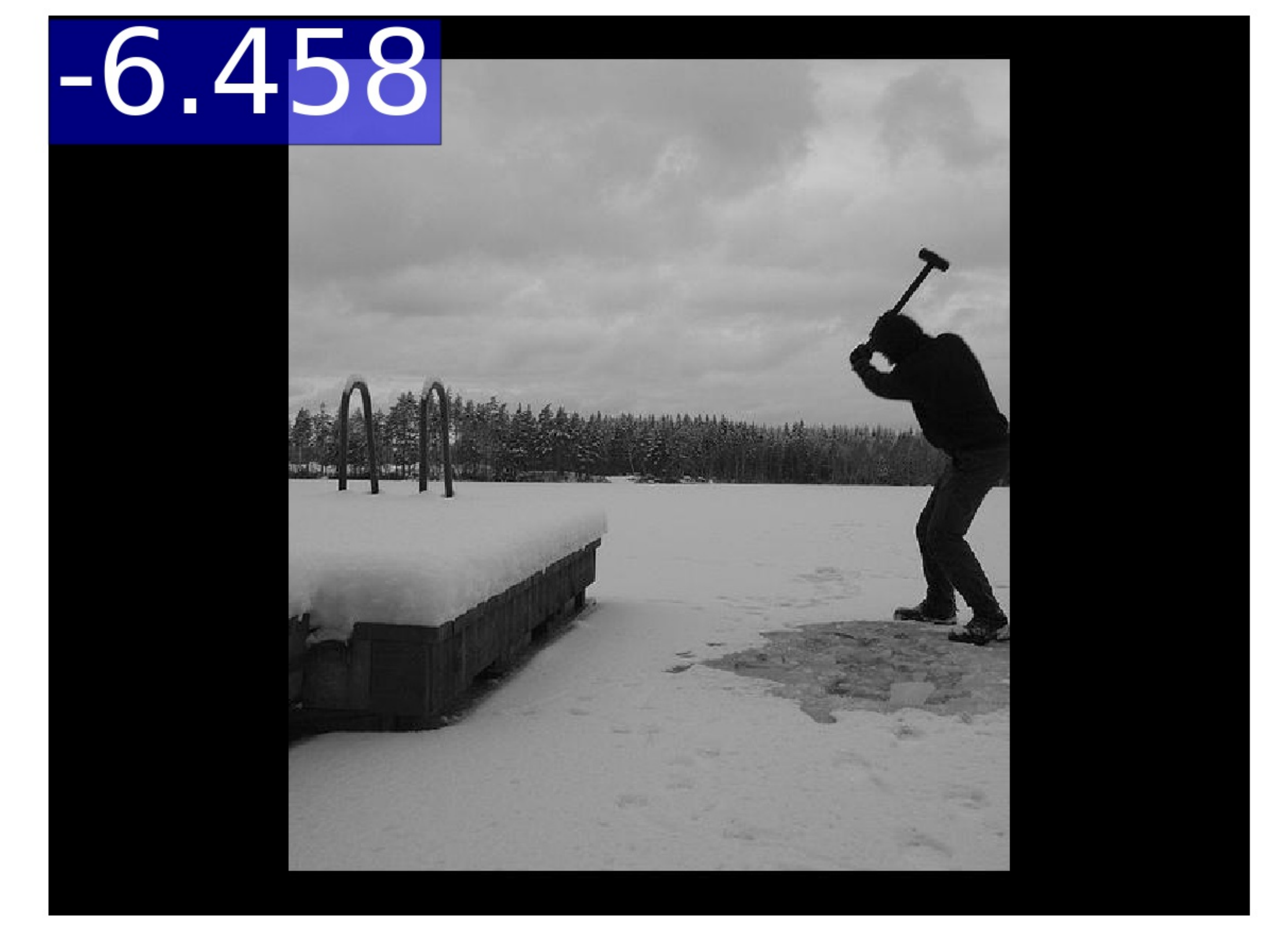}\\
\includegraphics[width=1\textwidth]{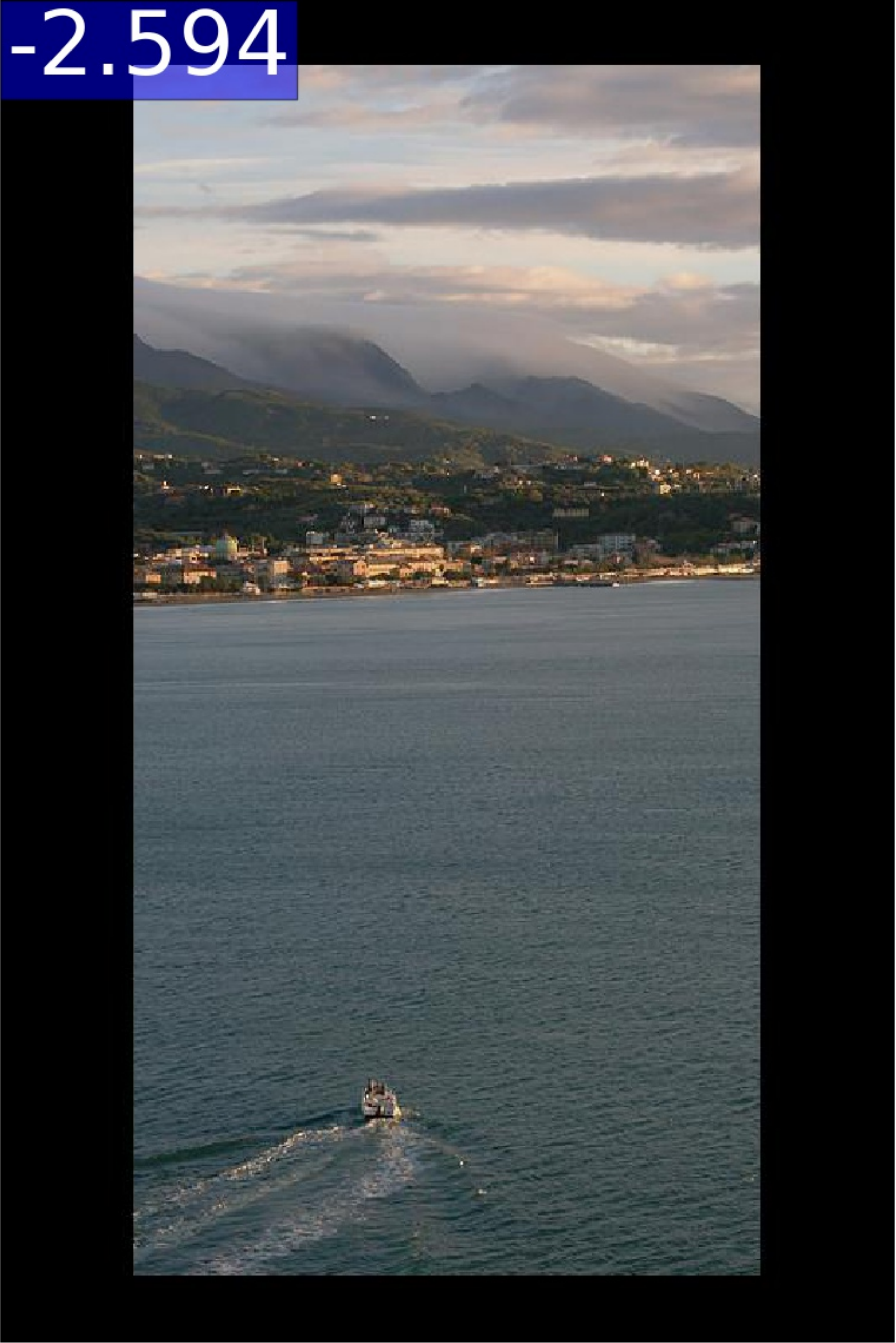}
\end{minipage}
}
\subfigure[A2-RL w/o LSTM]{
\centering
\begin{minipage}[b]{0.14\textwidth}
\includegraphics[width=1\textwidth]{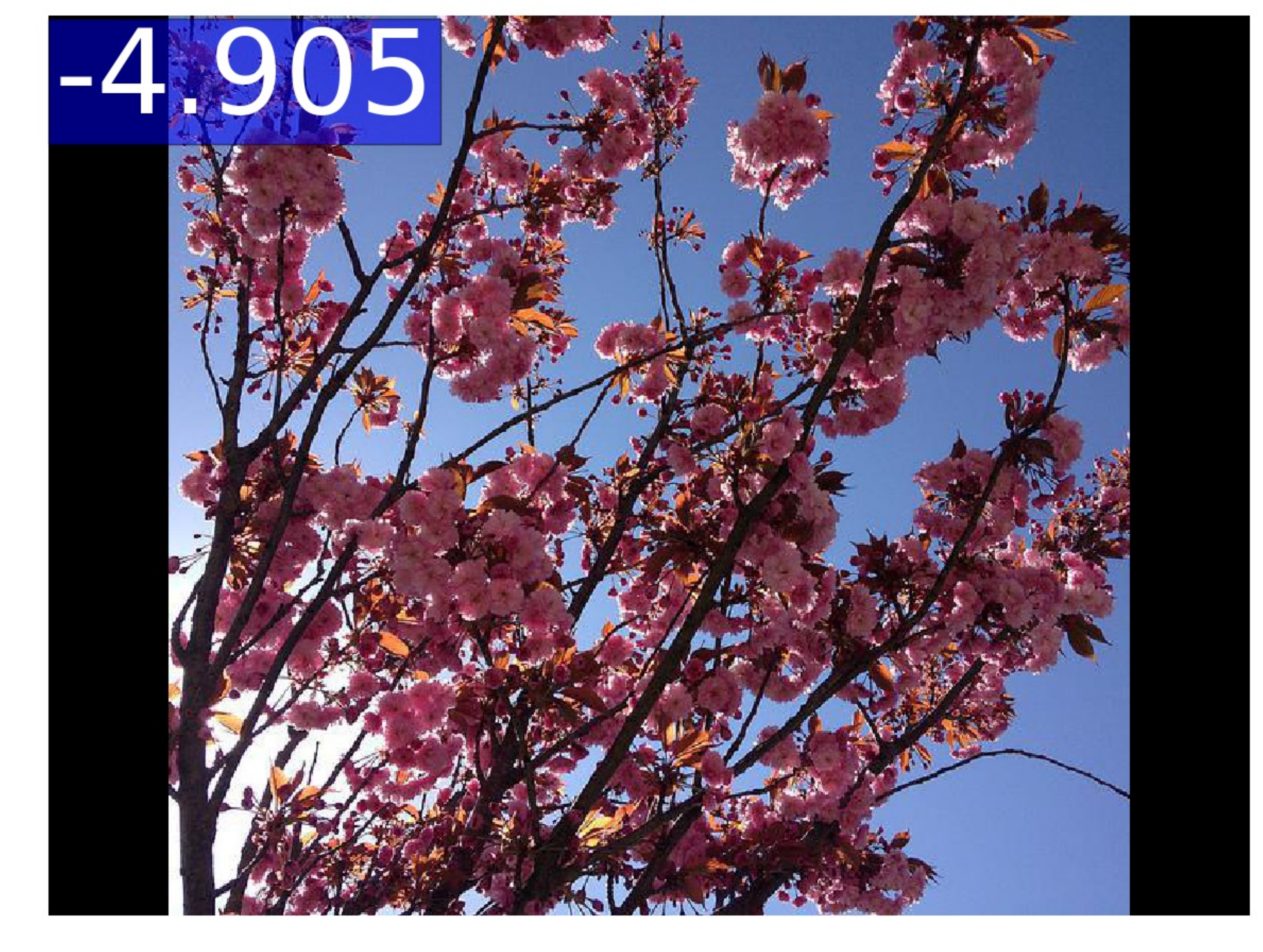} \\
\includegraphics[width=1\textwidth]{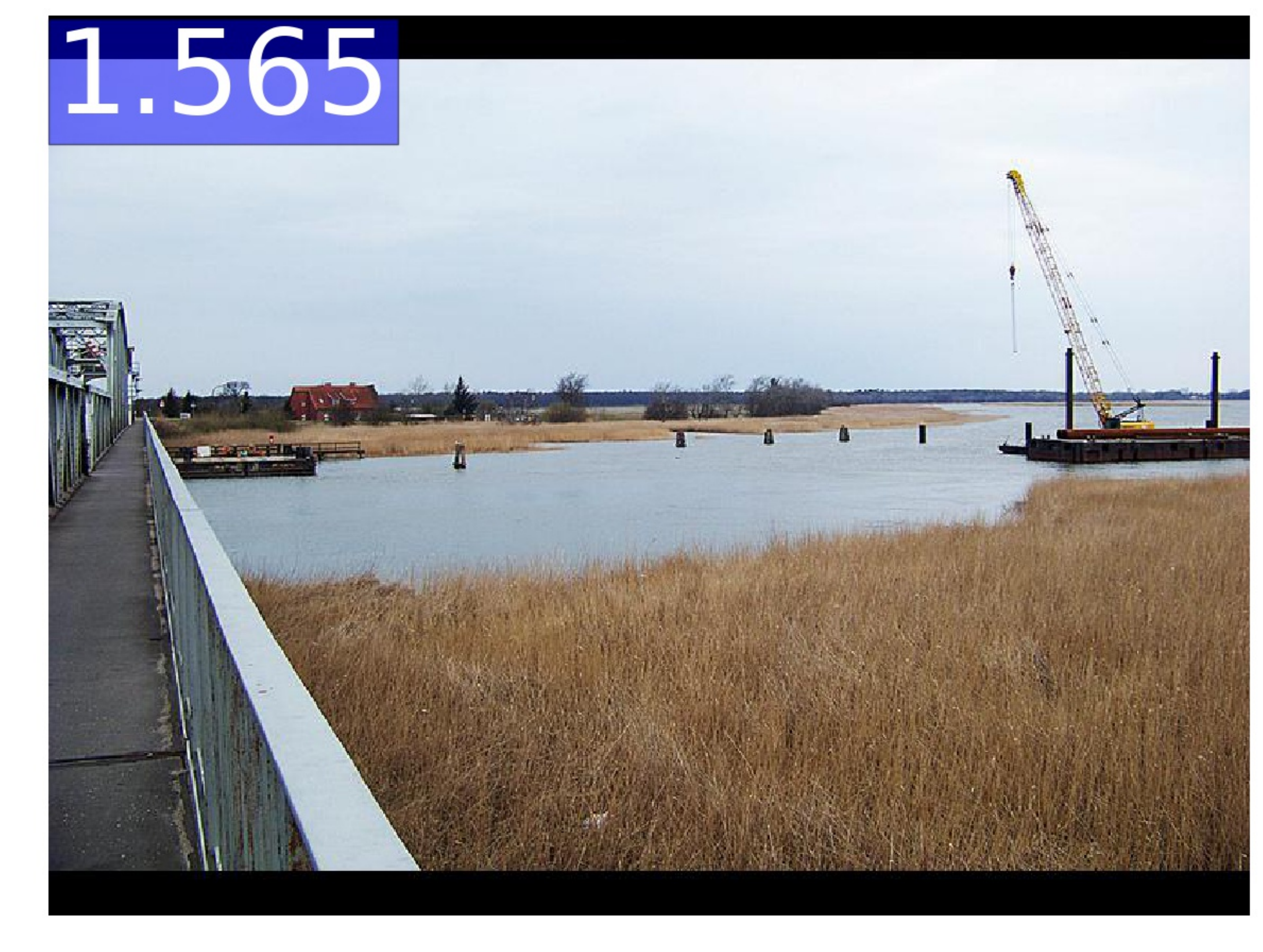} \\
\includegraphics[width=1\textwidth]{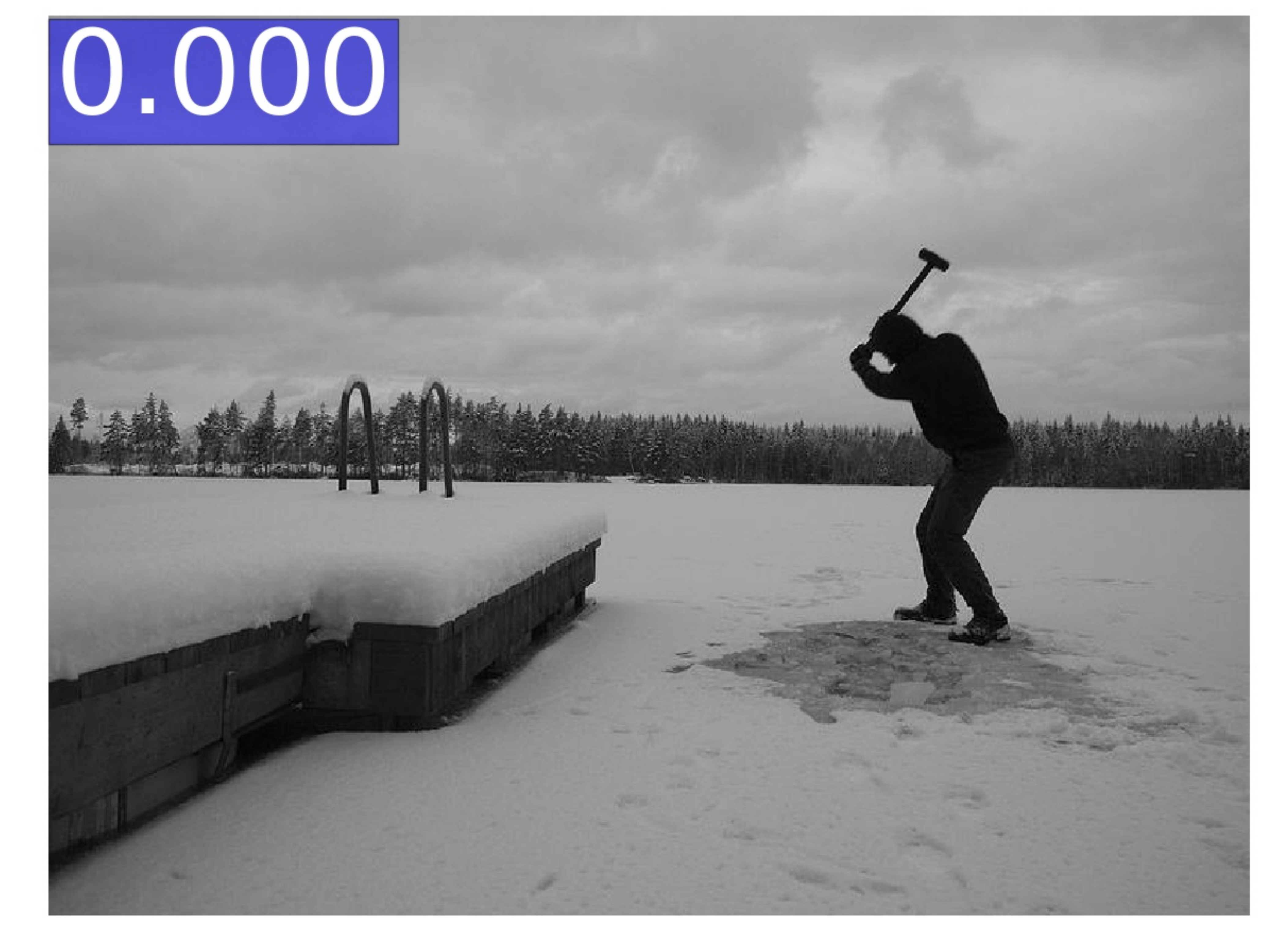}\\
\includegraphics[width=1\textwidth]{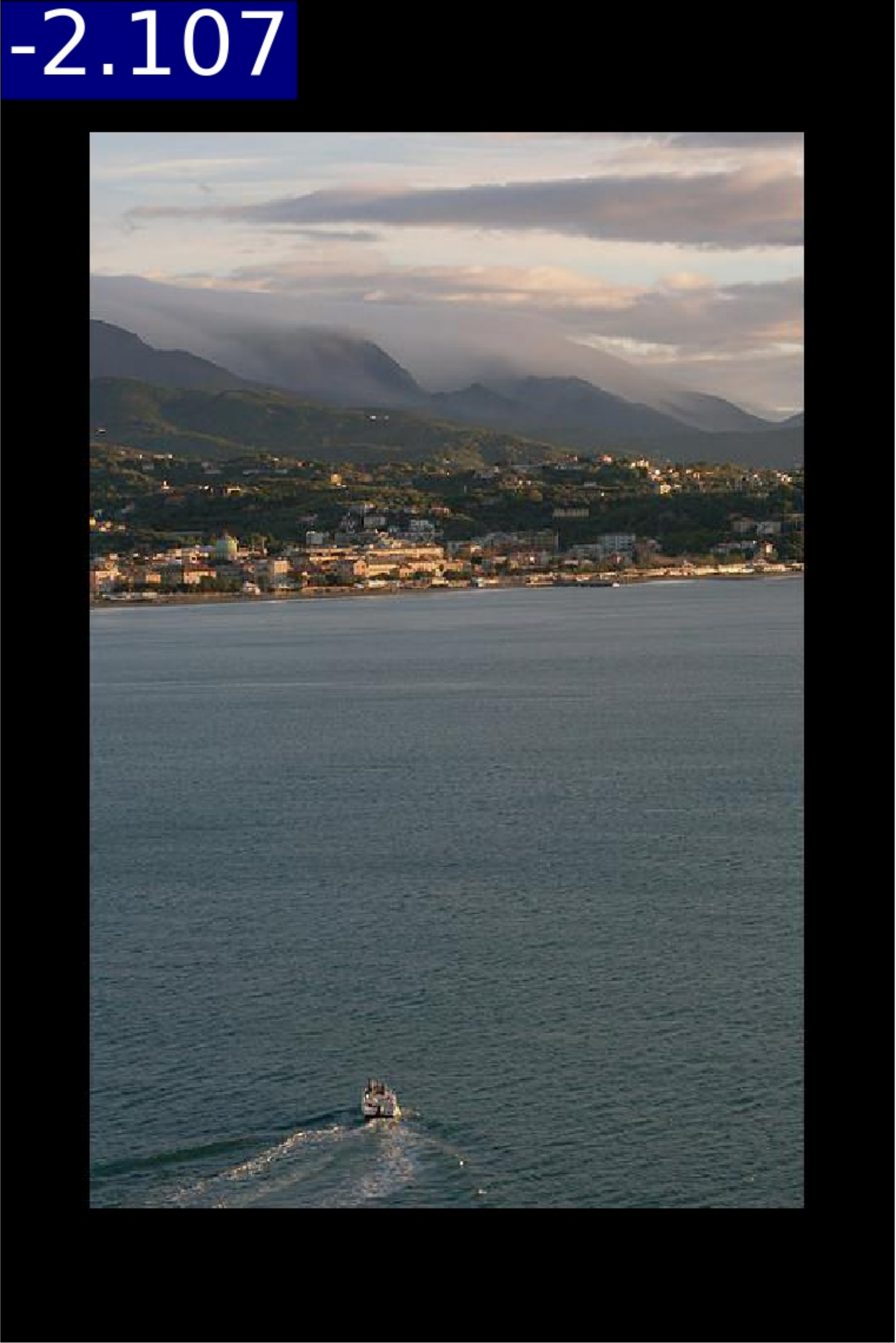}
\end{minipage}
}
\subfigure[A2-RL (Ours)]{
\centering
\begin{minipage}[b]{0.14\textwidth}
\includegraphics[width=1\textwidth]{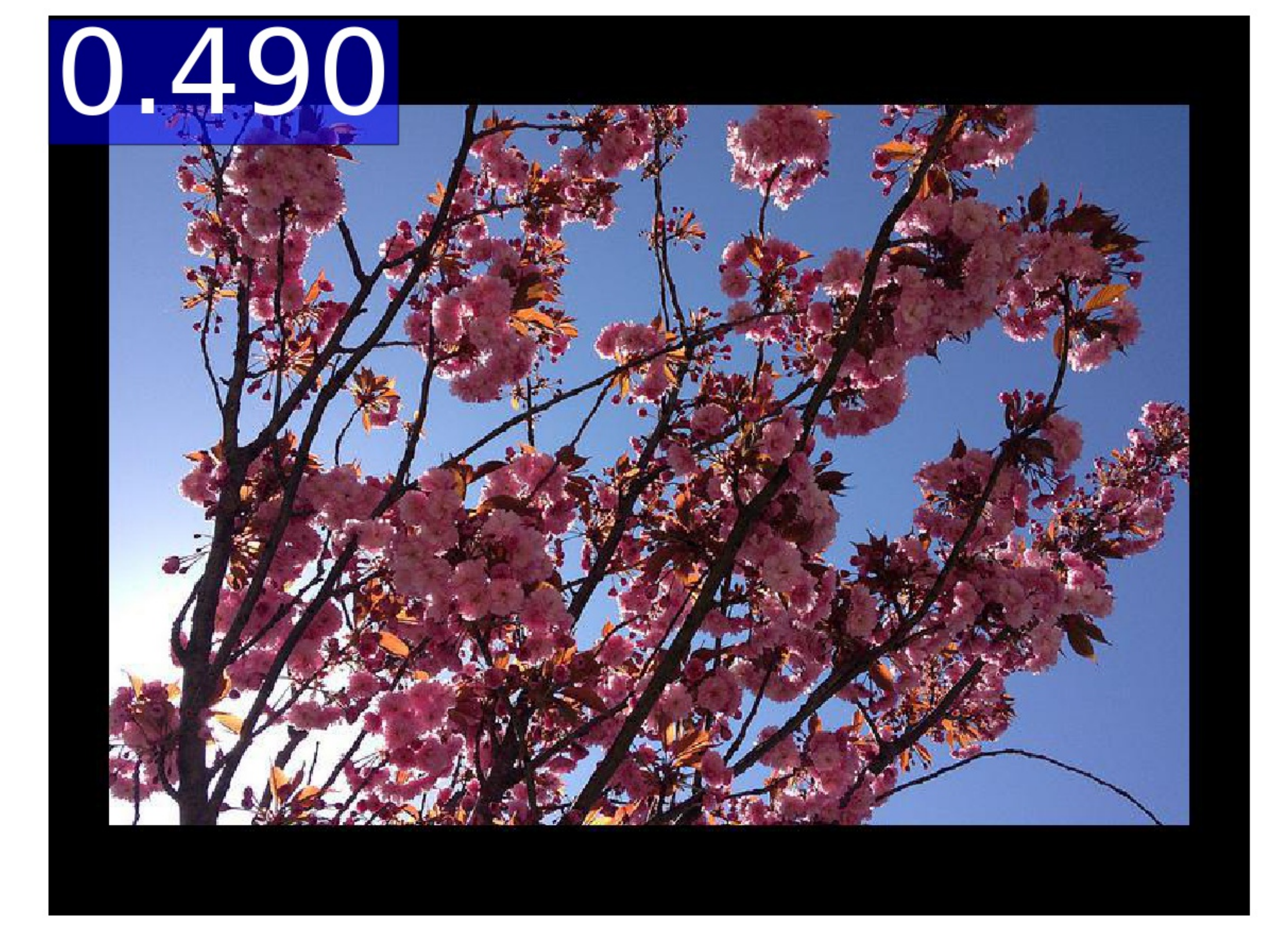} \\
\includegraphics[width=1\textwidth]{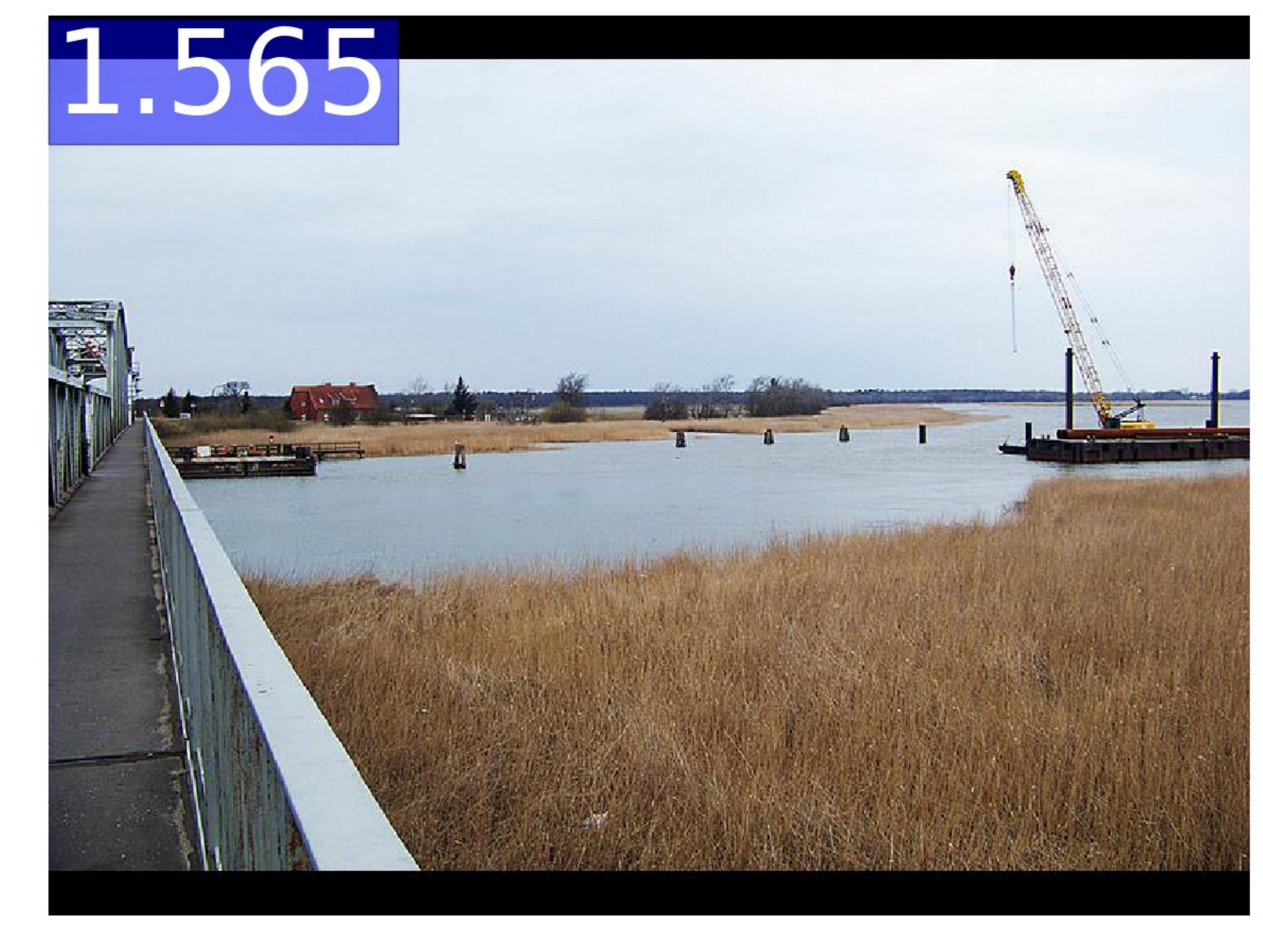} \\
\includegraphics[width=1\textwidth]{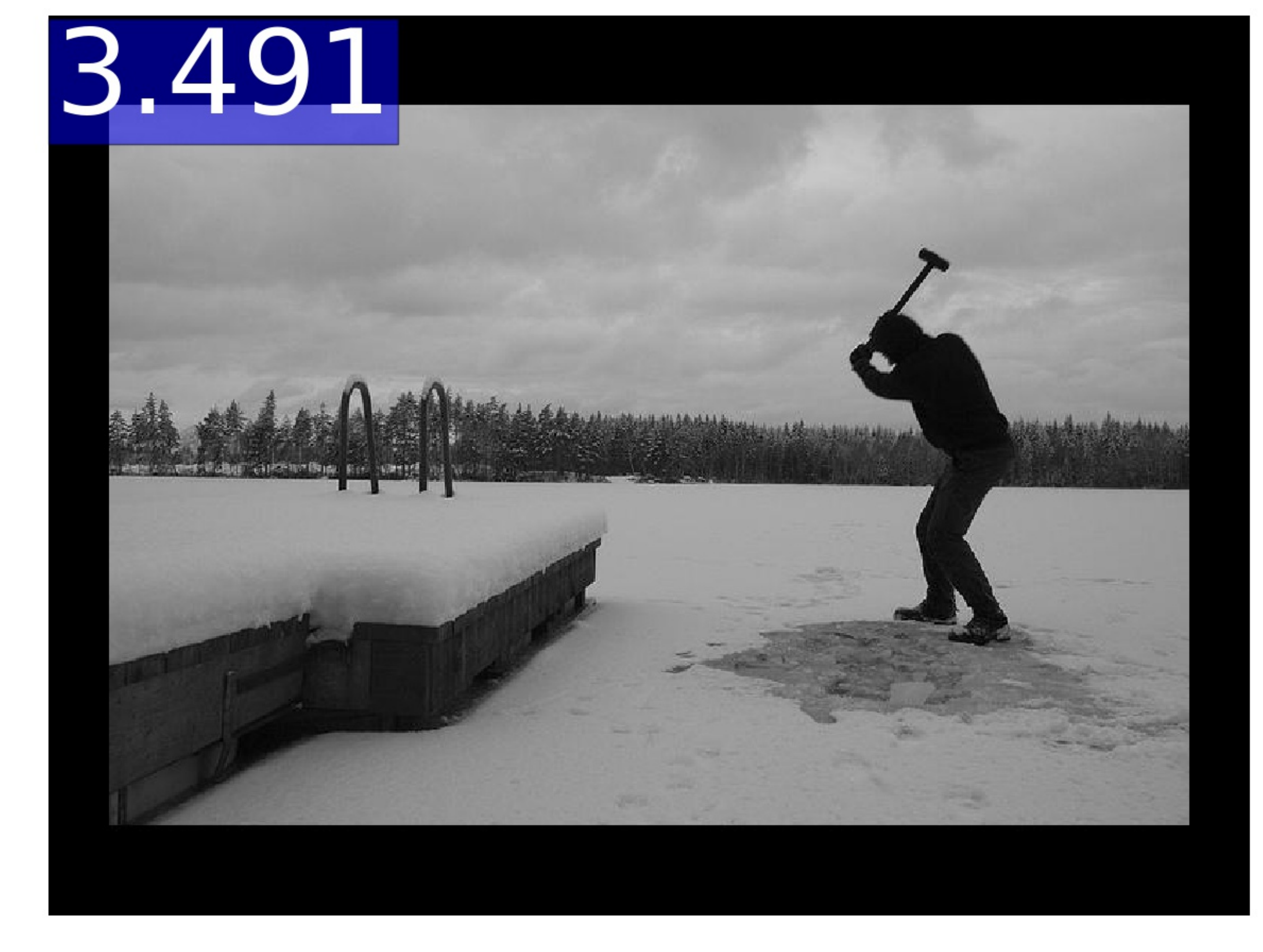}\\
\includegraphics[width=1\textwidth]{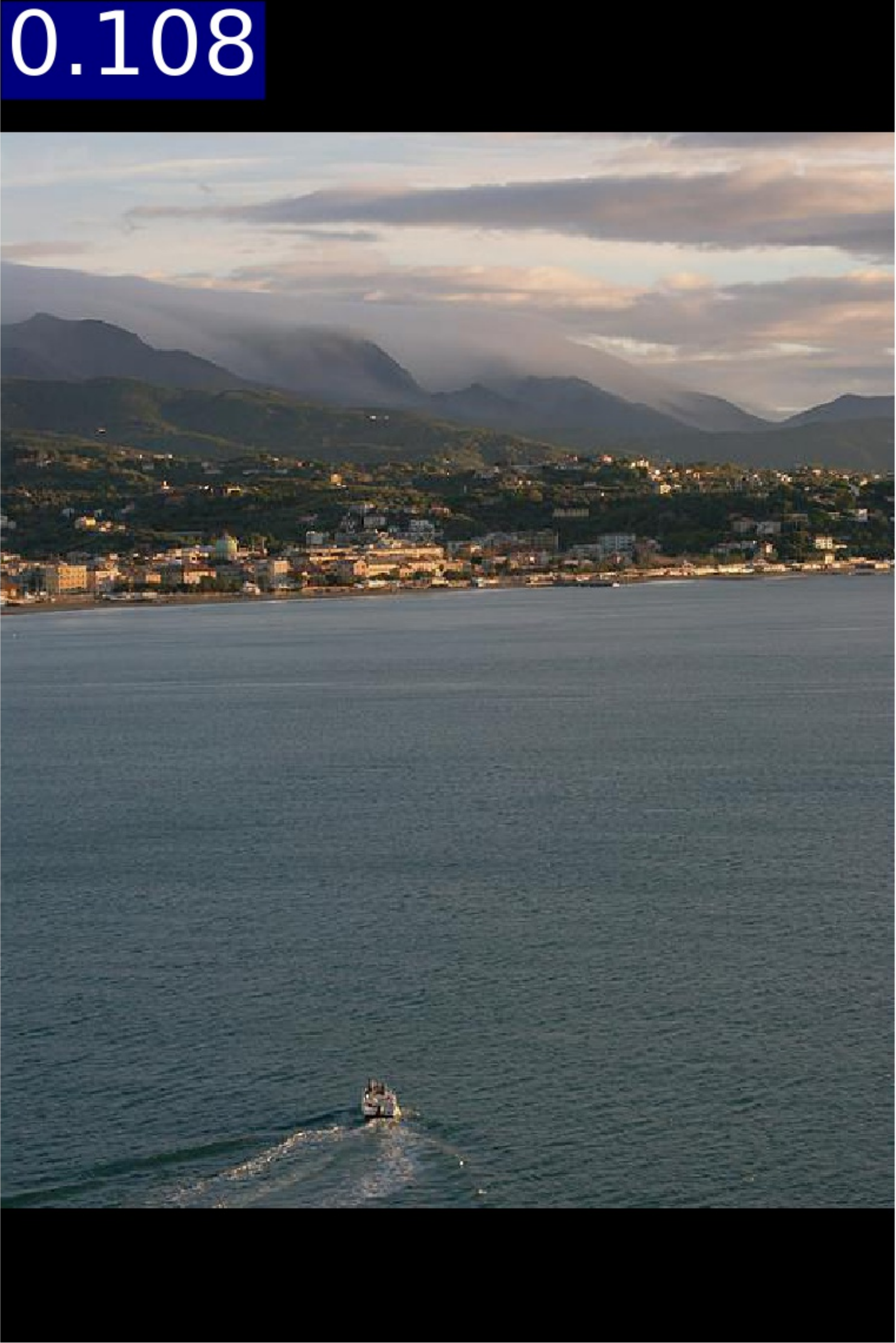}
\end{minipage}
}
\subfigure[Ground Truth]{
\centering
\begin{minipage}[b]{0.14\textwidth}
\includegraphics[width=1\textwidth]{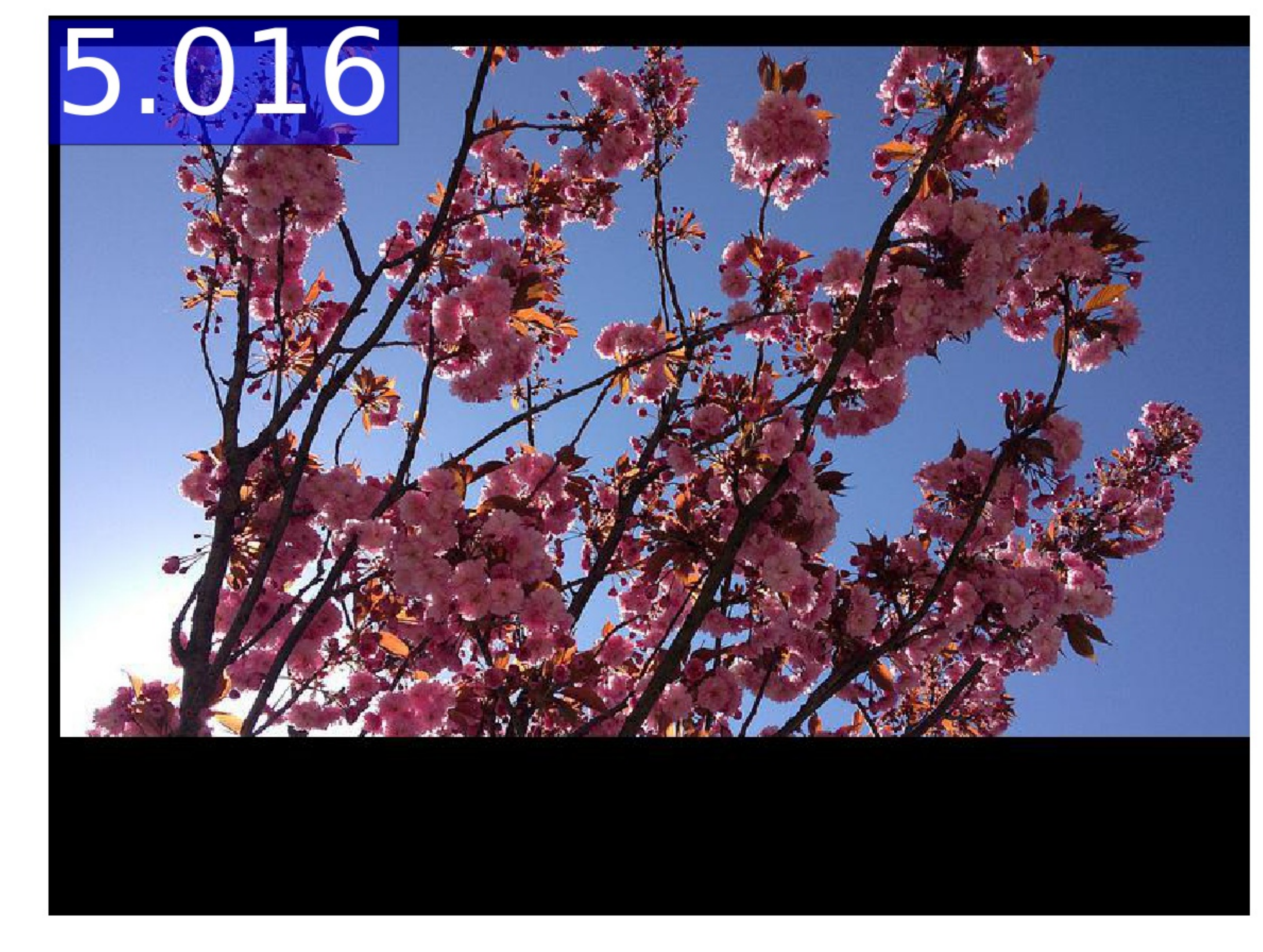} \\
\includegraphics[width=1\textwidth]{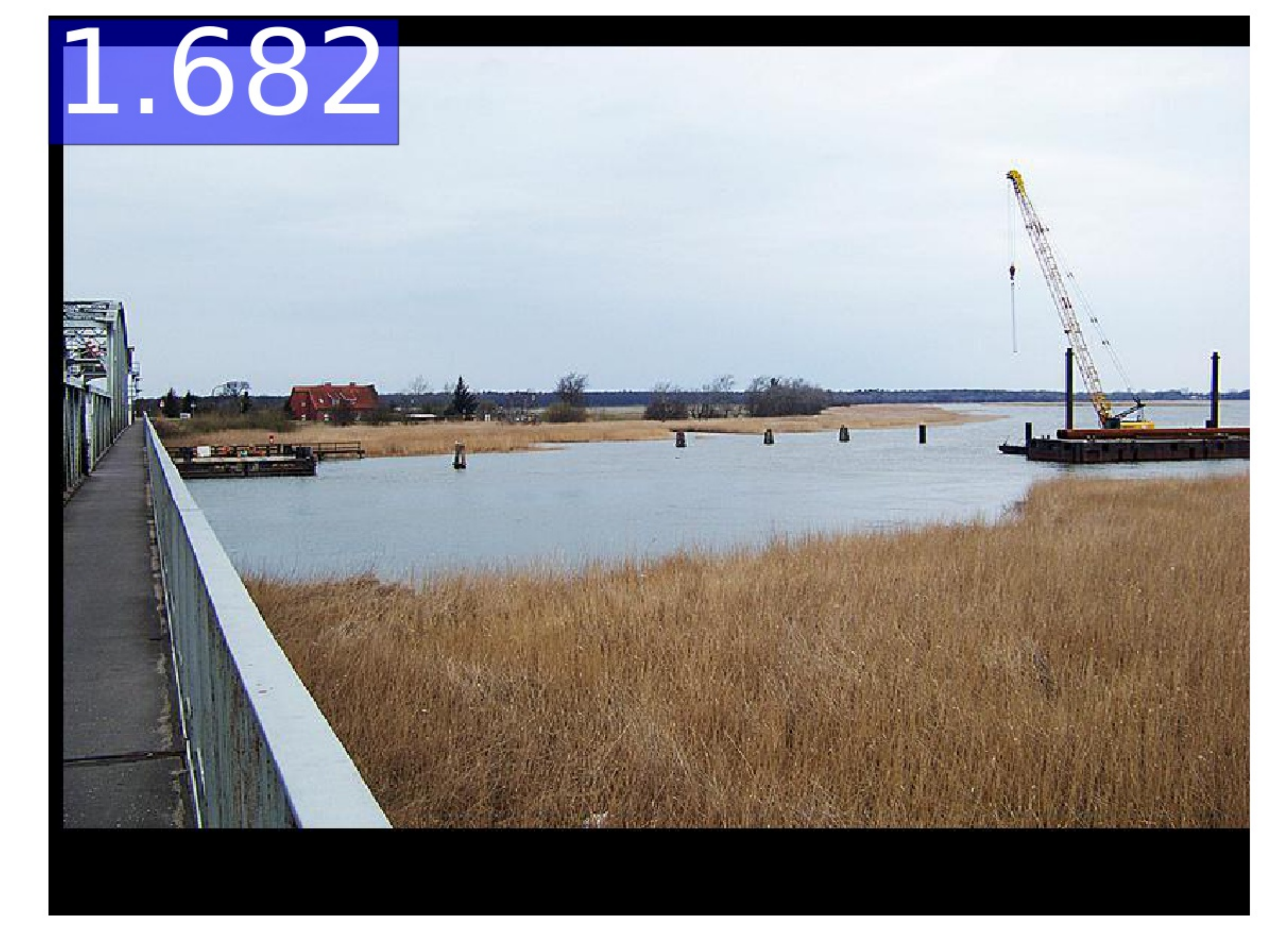} \\
\includegraphics[width=1\textwidth]{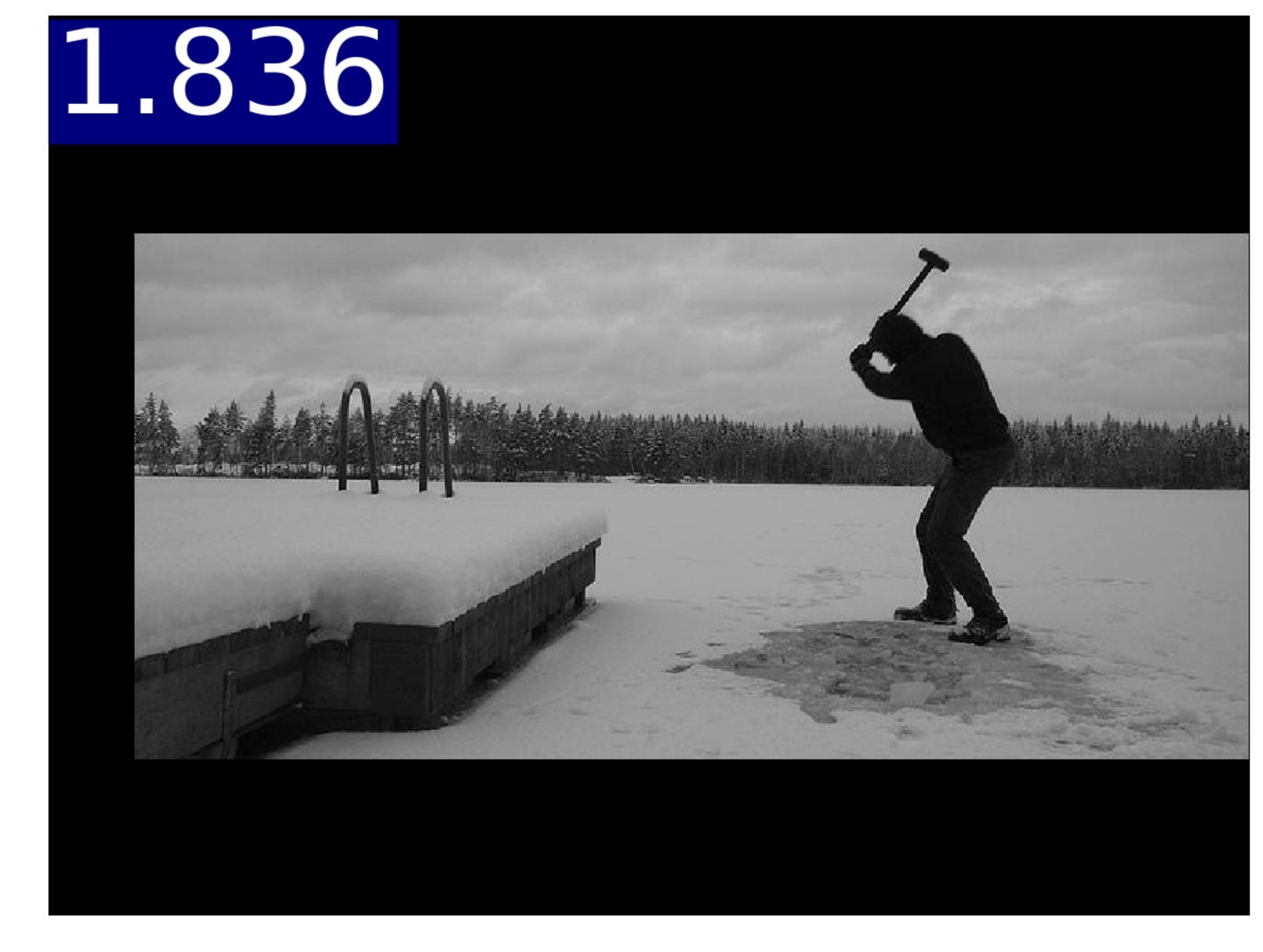} \\
\includegraphics[width=1\textwidth]{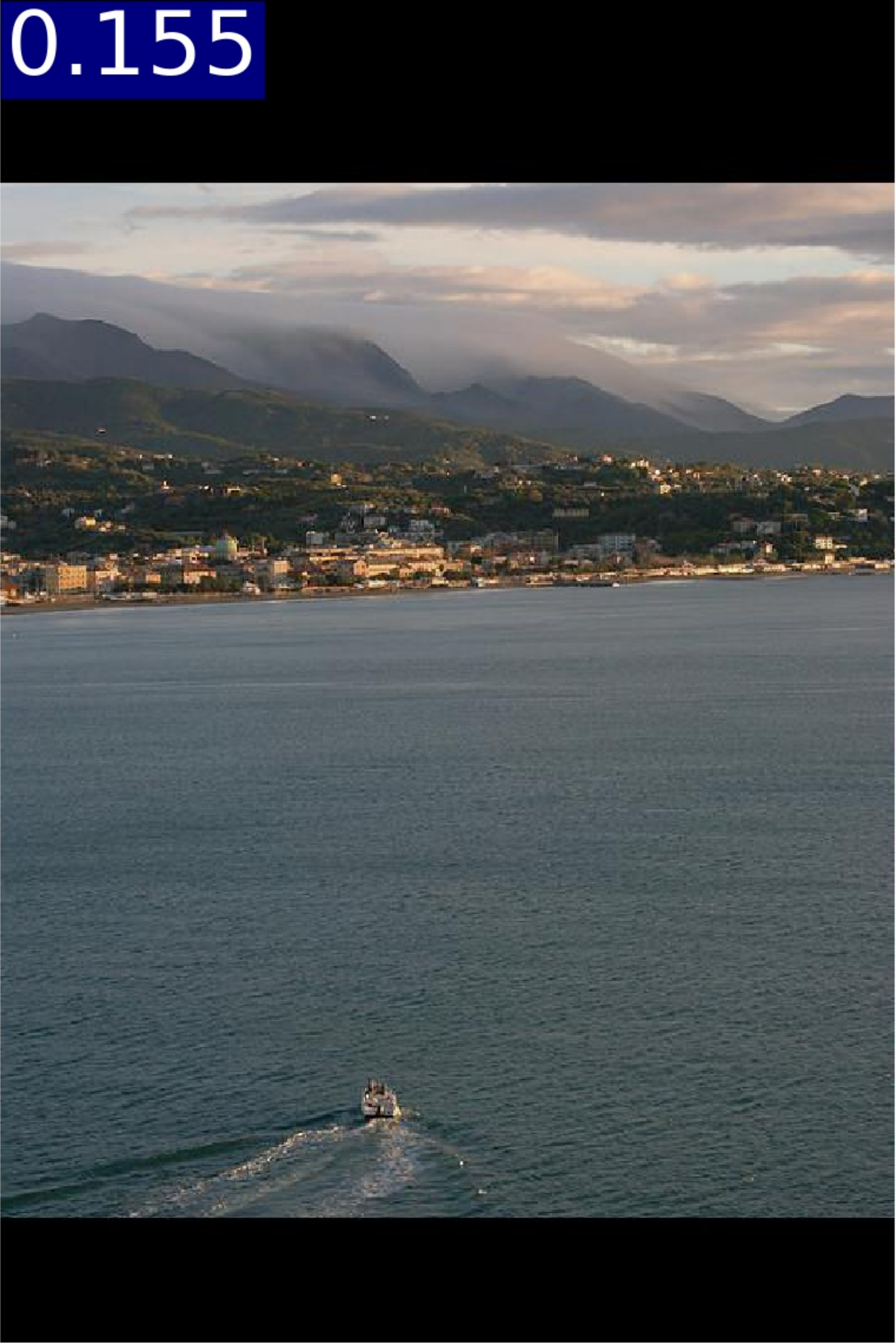}
\end{minipage}
}
\end{center}
\caption{Image cropping examples on FCD~\cite{chen2017quantitative}. The number in the upper left corner is the difference between the aesthetic scores of the cropped and original image, which is $s_{aes}(I_{crop})-s_{aes}(I_{original})$. The aesthetic score $s_{aes}(I)$ is used in the definition of the reward function (see Section~\ref{reward}). Best viewed in color.}
\label{fig:4}
\end{figure*}

\subsection{Qualitative Analysis}
We visualize how the agent works in our A2-RL model. We show the intermediate results of the cropping sequences, as well as the actions selected by the agent in each step. As shown in Figure~\ref{fig:3}, the agent takes the selected actions step by step to adjust the windows and chooses when to stop the process to get the best results.

We also show several cropping results of different methods on FCD~\cite{chen2017quantitative}. From Figure~\ref{fig:4}, we can find that the A2-RL model can find better cropping windows than other methods, which demonstrates the capabilities of our model in an intuitive way. Some results also show the importance of the limited aspect ratio and history experience.
\section{Conclusion}
In this paper, we formulated the aesthetic image cropping as a sequential decision-making process and firstly proposed a novel weakly supervised Aesthetics Aware Reinforcement Learning (A2-RL) model to address this problem. With the aesthetics aware reward and comprehensive state representation which includes both the current observation and historical experience, our A2-RL model learns good policies for automatic image cropping. The agent finished the cropping process within several or a dozens steps and got the cropping windows with almost arbitrary size. Experiments on several unseen cropping datasets showed that our model can achieve the state-of-the-art cropping accuracy with much fewer candidate windows and much less time.
\section*{Acknowledgement}
This work is funded by the National Key Research and Development Program of China (Grant 2016YFB1001004 and Grant 2016YFB1001005), the National Natural Science Foundation of China (Grant 61673375, Grant 61721004 and Grant 61403383) and the Projects of Chinese Academy of Sciences (Grant QYZDB-SSW-JSC006 and Grant 173211KYSB20160008).

{\small
\bibliographystyle{ieee}
\bibliography{A2RL}
}

\end{document}